%% file: main.tex
\documentclass[10pt,twocolumn,letterpaper]{article}
\usepackage{cvpr}


\definecolor{cvprblue}{rgb}{0.21,0.49,0.74}
\usepackage[pagebackref,breaklinks,colorlinks,allcolors=cvprblue]{hyperref}

\usepackage{bm}
\usepackage{multirow}
\usepackage{makecell}
\usepackage[table]{xcolor}
\usepackage{booktabs}
\usepackage{pifont}
\usepackage{xcolor}
\usepackage{enumitem}
\definecolor{darkred}{RGB}{130,0,0}
\definecolor{darkgreen}{RGB}{0,130,0}
\definecolor{darkblue}{RGB}{191, 255, 255}

\title{Restore-R1: Efficient Image Restoration Agents via Reinforcement Learning with Multimodal LLM Perceptual Feedback}

\author{Jianglin~Lu$^{1,2}\thanks{Corresponding author. Work done during an internship at Amazon.}$, Yuanwei~Wu$^{1}$, Ziyi~Zhao$^{1}$, Hongcheng~Wang$^{1}$, Felix~Jimenez$^{1}$, Abrar~Majeedi$^{1,3}$, Yun~Fu$^{2}$\\
$^{1}$Amazon, $^{2}$Northeastern University, $^{3}$University of Wisconsin-Madison \\
{\tt\small jianglinlu@outlook.com, \{ywuvlms,zhaziyi,hongchw\}@amazon.com, yunfu@ece.neu.edu}
}

\begin{document}
\maketitle
\input{sec/0_abstract}    
\input{sec/1_intro}
\input{sec/2_formatting}
\input{sec/3_methodology}

\input{sec/4_experiments}
\input{sec/5_conclusion}

{
    \small
    \bibliographystyle{ieeenat_fullname}
    \bibliography{main}
}

 
\input{sec/X_suppl}

\end{document}

%% file: sec/0_abstract.tex
\begin{abstract}
Complex image restoration aims to recover high-quality images from inputs affected by multiple  degradations such as blur, noise, rain, and compression artifacts.
Recent restoration agents, powered by vision-language models and large language models, offer promising restoration capabilities but suffer from significant efficiency bottlenecks due to reflection, rollback, and iterative tool searching.
Moreover, their performance heavily depends on degradation recognition models that require extensive annotations for training, limiting their applicability in label-free environments.
To address these limitations, we propose a policy optimization-based restoration framework that learns an lightweight agent to determine tool-calling sequences.
The agent operates in a sequential decision process, selecting the most appropriate restoration operation at each step to maximize final image quality.
To enable training within label-free environments, we introduce a novel reward mechanism driven by multimodal large language models, which act as human-aligned evaluator and provide perceptual feedback for policy improvement. 
Once trained, our agent executes a deterministic restoration plans without redundant tool invocations, significantly accelerating inference while maintaining high restoration quality.
Extensive experiments  show that despite using no supervision, our method matches SOTA performance on full-reference metrics and surpasses existing approaches on no-reference metrics across diverse degradation scenarios.
\end{abstract}

%% file: sec/1_intro.tex
\section{Introduction}
\label{sec:intro}

\begin{figure}
\centering
\includegraphics[scale=0.58,trim=169 86 145 82,clip]{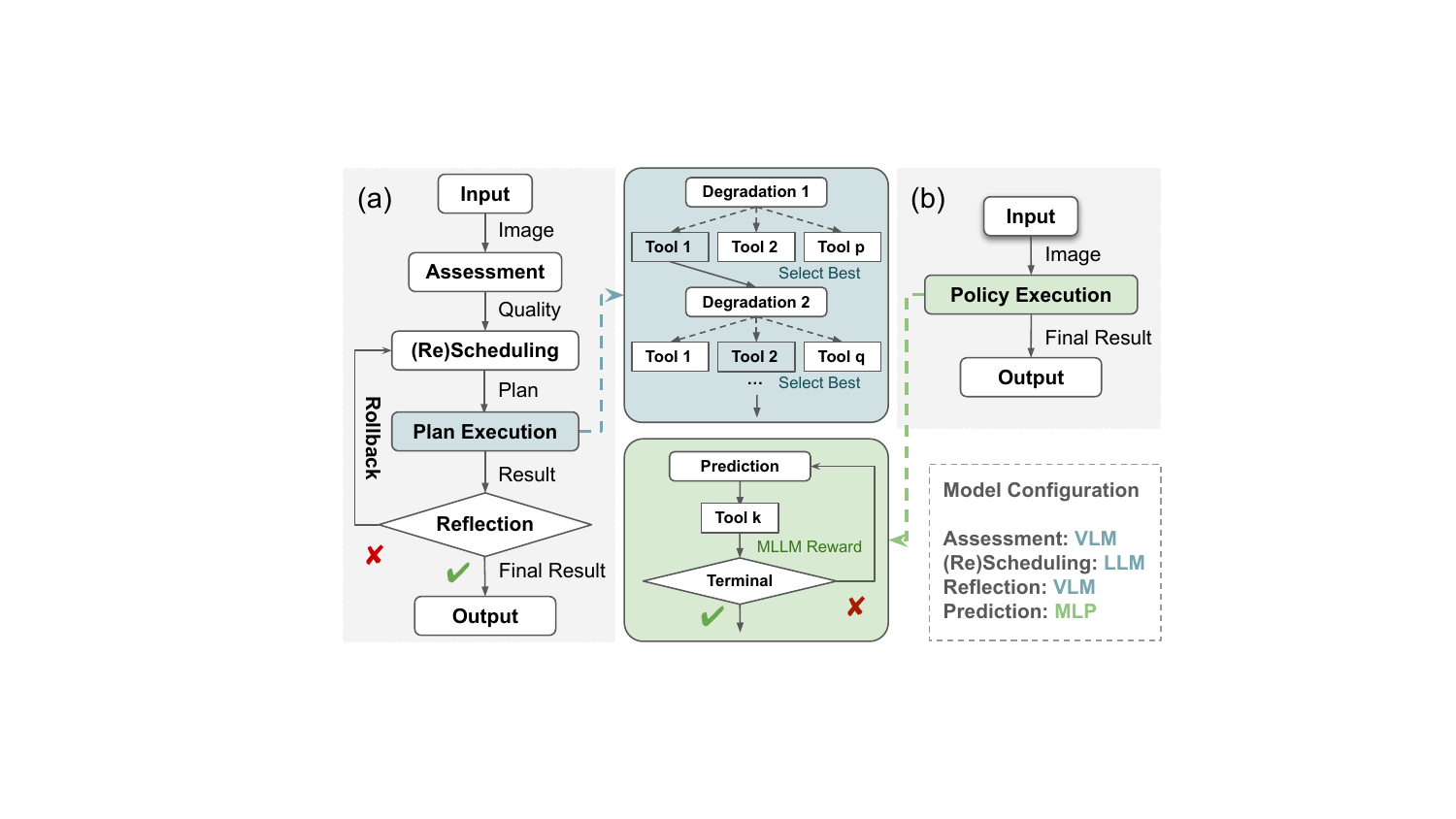}
\caption{(a) Existing restoration agents \cite{RestoreAgent, agenticir, hybridagent} typically consist of assessment, scheduling, execution, reflection, and rollback, using VLMs for degradation recognition and LLMs for plan making; (b) Our Restore-R1 agent determines the tool-calling sequence via
a single policy execution, avoids the need for iterative trial-and-error, and generalizes to label-free environments.}
\label{figxxx}
\end{figure}

Image restoration aims to reconstruct high-quality images from degraded observations, where the underlying corruption process is often unknown and involves complex interactions \cite{wang2025outlier, unsupervised1, wang2023ift}.
In real-world scenarios, low-quality images rarely suffer from a single type of distortion; instead, they typically exhibit a mixture of blur, rain streaks, noise, haze, low-light conditions, and compression artifacts, among others \cite{Wang2021, he2023weakly, LiFlexible, FuCamouflaged, lu2026seeing}.
This setting, referred to as \textit{complex image restoration (CIR)} \cite{RestoreAgent, agenticir, multiAgent, hybridagent}, is considerably more challenging than classical single-degradation restoration due to compound, nonlinear, and spatially varying corruption patterns. CIR further introduces fundamental difficulties, including ambiguous degradation sources, the absence of paired clean supervision, and degradation-stacking effects that amplify visual deterioration.

\begin{table*}[!tb]
\centering
\small
\setlength{\tabcolsep}{3.5pt}
\caption{Comparison with existing image restoration agents, where \ding{51} indicates the feature is required and \ding{55} indicates that it is not.}
\begin{tabular}{l|llllcc}
\toprule
{Agentic Method} & \# of Params & Restoration Planner & Ground-Truth & Recognition Model    & Reflection & Rollback \\
\midrule
RestoreAgent \cite{RestoreAgent}  & 8B & Llava-Llama3 \cite{touvron2023llama}  & Optimal Sequence   & Llava-Llama3 \cite{touvron2023llama}  & \ding{51}  & \ding{51}  \\
AgenticIR  \cite{agenticir} & 7B  & GPT-4 \cite{achiam2023gpt} & Image+Label+Level & DepictQA \cite{DepictQA}  & \ding{51}  & \ding{51} \\
MAIR \cite{multiAgent}    & 7B  & GPT-4o \cite{achiam2023gpt}  & Image+Label+Level & DepictQA \cite{DepictQA}  &  \ding{51} & \ding{55} \\
Q-Agent  \cite{QAgent}     & 7B  & Greedy Strategy  & Image+Label+Level  & Qwen2-VL \cite{wang2024qwen2} & \ding{51} & \ding{55} \\
HybridAgent \cite{hybridagent} & 8.2B & Llama3.2-Instruct \cite{dubey2024llama}    & Image+Label  & Co-instruct \cite{wu2024towards}  &  \ding{51} & \ding{51} \\
Ours   & 0.28B & Image Encoder + MLP  & \qquad \ding{55} & \qquad\ding{55} & \ding{55} & \ding{55}  \\
\bottomrule
\end{tabular}
\label{agents_comparison}
\end{table*}

To address the CIR problem, a variety of methods have emerged in recent years. A prominent direction focuses on all-in-one restoration, which handles heterogeneous and often unknown degradations with unified restoration models \cite{airnet, kong2024towards, park2023all, Autodir, yang2024all}.
Early solutions primarily adopted unified CNN architectures to learn general low-level image priors, such as MPRNet \cite{MPRNet} and MIRNet \cite{MIRNet}. More recent methods leverage Transformer-based architectures, including Restormer \cite{Restormer} and Uformer \cite{Uformer}, to better capture long-range dependencies beneficial for joint restoration. 
To further improve adaptability to heterogeneous distortions, various conditioning and modulation strategies have been introduced, such as degradation embeddings (e.g., AirNet \cite{airnet}), prompt-based conditioning (e.g., PromptIR \cite{Promptir}, InstrucIR \cite{Instructir}), dynamic expert routing  (e.g., AMIR \cite{yang2024all}), CLIP-driven control (e.g., DA-CLIP \cite{daclip}), and generative priors (e.g., AutoDIR \cite{Autodir}). 
While providing a general solution, all-in-one models often struggle with complex real-world degradation mixtures, exhibiting limited flexibility and sub-optimal performance due to the need to fit broad degradation distributions \cite{RestoreAgent, agenticir, Wang2021, wang2025outlier}.

With the rapid progress of intelligent agents across multiple domains, agent-based image restoration has recently emerged as a compelling alternative paradigm. Inspired by human-like decision-making, these methods leverage large language models (LLMs) \cite{achiam2023gpt, touvron2023llama, dubey2024llama, lu2026seeing} and vision-language models (VLMs) \cite{DepictQA, wang2024qwen2, zhang2025linkedout, zhang2025vqtoken} to autonomously recognize  image degradations, plan restoration pipelines, invoke appropriate restoration tools, and progressively refine outputs toward high-quality results.
The state-of-the-art (SOTA) approaches include RestoreAgent \cite{RestoreAgent}, AgenticIR \cite{agenticir}, MAIR \cite{multiAgent}, Q-Agent \cite{QAgent}, and HybridAgent \cite{hybridagent} (see Table \ref{agents_comparison} for a detailed comparison among them).
While demonstrating promising performance, these agent-driven frameworks suffer from substantial efficiency limitations.
Determining the optimal sequence of restoration tools, especially involving iterative reflection, trial-and-error, and tool re-scheduling (as shown in Figure~\ref{figxxx}), introduces heavy computational overhead and prolonged inference time. This process becomes particularly more time-consuming and suboptimal as the degradation patterns become more diverse and complex.
Moreover, most existing agents rely on a recognition model to identify degradations for tool selection, which requires extensive annotations for training, such as degradation labels \cite{agenticir, multiAgent} or optimal tool-calling trajectories \cite{RestoreAgent}. Such supervision dependencies restrict their universality and deployment in real-world, label-free environments.

To overcome these challenges, we propose a policy optimization-based image restoration framework that learns a lightweight agent, named Restore-R1, to select tools and determine their execution order, aiming to maximize overall restoration quality.
Our framework adopts an actor–critic architecture, where the actor network sequentially selects most appropriate restoration tools based on the observed image state, while the critic network estimates the expected return and provides feedback guidance to assess whether the chosen actions outperform the expected behavior. This formulation enables our agent to progressively refine degraded inputs through a sequence of learned restoration decisions.
To handle the label-free environments, we introduce a perceptual reward mechanism powered by multimodal large language models (MLLMs) \cite{dong2026co, lu2025the, kong2025token, dong2026refadv, lu2025representation}, which serve as a perception evaluator to provide human-aligned feedback on restored outputs.
The actor is then optimized based on these reward signals to encourage operation sequences that yield higher cumulative gains. 
To ensure stable policy learning and avoid abrupt policy shifts, we adopt the clipping strategies during training as in \cite{PPO, GRPO}.
During inference, our agent executes its learned policy in a single forward pass to deterministically produce restoration sequences, eliminating iterative trial-and-error loops and redundant tool invocations typically required in prior agent-based solutions \cite{agenticir, multiAgent, RestoreAgent} (as shown in Figure \ref{figxxx}).
In summary, our main contributions are as follows:
\begin{itemize}
    \item To the best of our knowledge, we introduce the first label-free agentic framework for complex image restoration, requiring no ground-truth images, degradation labels, or optimal sequence guidance.

    \item We design a lightweight agent that directly determines tool-calling sequences  through a single policy execution, eliminating expensive trial-and-error loops used in prior agentic approaches \cite{RestoreAgent, agenticir, multiAgent}.

    \item We propose a novel and general reward mechanism that leverages MLLMs to provide human-like perceptual feedback for action evaluation and policy optimization. 
    
    \item Extensive experiments show that our agent matches supervised SOTA methods on full-reference metrics and outperforms them significantly on no-reference metrics.
    
\end{itemize}

%% file: sec/2_formatting.tex
\section{Related Work}
\label{related_work}

\subsection{Complex Image Restoration}
\label{section21}

\begin{figure*}
\centering
\includegraphics[scale=0.79,trim=56 124 60 107,clip]{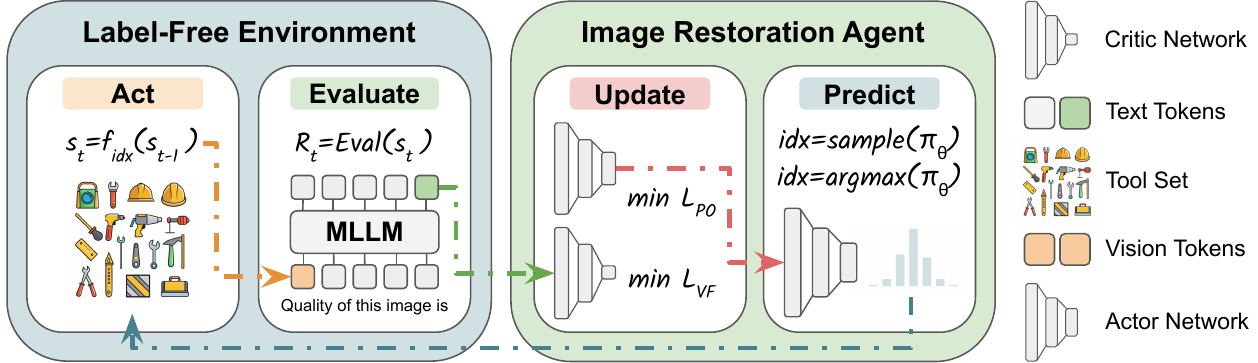}
\caption{Framework overview. The restoration agent predicts the next action based on the current input status (sampling actions during training while selecting the highest-probability action during inference, see Sec. \ref{ira_}). 
The environment executes the chosen action, evaluates the restored output with an MLLM, and returns a feedback signal to update the agent's policy (see Sec. \ref{LF_ENV}). Through this iterative interaction, the agent progressively refines its decision-making policy without ground-truth supervision.}
\label{main4_fig}
\end{figure*}

\textbf{All-in-One Image Restoration}. 
All-in-one methods \cite{airnet, kong2024towards, park2023all, Autodir} aim to handle diverse and often unknown degradations within a single unified framework. 
For example, AirNet \cite{airnet} learns a unified backbone capable of adapting to various corruption types.
PromptIR \cite{Promptir}
introduces employs prompt-based conditioning to steer restoration and dynamically handle different degradations.
DA-CLIP \cite{daclip}
leverages vision-language priors and natural-language instructions to flexibly control restoration behavior, and InstrucIR \cite{Instructir} similarly aligns restoration outputs with human instructions.
AutoDIR \cite{Autodir}
leverages latent diffusion models to automatically infer and correct degradations without explicit user instructions. 
Although these approaches provide a general solution, all-in-one models often struggle with complex real-world degradation mixtures, exhibiting limited flexibility and sub-optimal performance due to the need to fit broad degradation distributions \cite{wang2023ift, hybridagent, agenticir, wang2025outlier}.

\textbf{Agentic Image Restoration}.
Agent-based approaches treat restoration as a process of iterative decision making, where an intelligent agent analyzes the input image, identifies degradation types, and composes a sequence of specialized restoration tools.
For example, RestoreAgent \cite{RestoreAgent} fine-tunes a VLM to directly generate execution plans, while AgenticIR \cite{agenticir} integrates VLM-based quality analysis with LLM-based planning following a human-inspired pipeline. 
Q-Agent \cite{QAgent} improves efficiency via a greedy planner driven, while MAIR \cite{multiAgent} categorizes degradations into scene, imaging, and compression types, and reverses them in an inverse order.
HybridAgent \cite{hybridagent} designs routing strategy to balance accuracy and efficiency.
Despite notable progress, existing agent-based systems often suffer from high inference overhead due to extensive search, and their effectiveness heavily relies on auxiliary recognition models trained with ground-truth sequences or degradation labels, which limits their practicality and generalization in label-free real-world scenarios. 
Table~\ref{agents_comparison} presents a comprehensive comparison between these approaches.

\subsection{Reinforcement Learning}
Reinforcement learning (RL) \cite{mnih2015human, silver2016mastering, silver2017mastering, DQN, ouyang2022training, wang2025otc} frames decision making as an interactive optimization problem, where an agent refines its actions through feedback from the environment to achieve maximal cumulative benefit.
Most RL algorithms fall into two categories: value-based and policy-based.
Value-based techniques estimate action-dependent returns through a value function and induce a decision strategy by favoring actions with higher predicted returns.
Representative examples include Q-learning \cite{QLearning} and its deep variant DQN \cite{DQN}. 
In contrast, policy-based approaches explicitly optimize the policy parameters, typically via gradient-based optimization, to maximize expected rewards. Representative examples include REINFORCE \cite{williams1992simple}, TRPO \cite{TRPO}, PPO \cite{PPO}, RLHF \cite{ouyang2022training} and the recently proposed GRPO \cite{GRPO}. In this work, we utilize RL to determine optimal tool-execution strategies for complex image restoration.
Unlike prior RL-based approaches \cite{park2018distort, yu2018crafting, qiao2025realsr, DSPO}, which often rely on predefined task labels, heuristic rules, or supervised sequence annotation, we learn a policy that autonomously selects both restoration tools and their execution order in a label-free environment, guided solely by perceptual feedback from MLLMs. 

%% file: sec/3_methodology.tex
\section{Restore-R1 Framework}
\label{methodology}

\subsection{Problem Definition}
In complex image restoration, we are given a library of $n$ restoration tools $\mathcal{F}=\{f_1, f_2, \ldots, f_n\}$, where each tool can handle one or more specific degradations.  Given a degraded input $x_{L}$, the objective is to recover a clean image $\widetilde x_{H}$ by executing a sequence of restoration tools:
\begin{equation}
\widetilde x_{H} = (f_k \circ f_{k-1}  \circ \cdots \circ f_0) (x_{L}),
\end{equation}
where $f_t$ denotes the tool selected at step $t$.
The core challenge lies in identifying the optimal tool-calling sequence:
\begin{equation}
\tau^{*}:=f_k^* \circ f_{k-1}^*  \circ \ldots \circ f_0^*  
\end{equation}
that minimizes the discrepancy between $\widetilde x_{H}$ and the clean image $x_{H}$.
Existing agent-based methods either (i) finetune a large vision-language model to directly predict the tool sequence \cite{RestoreAgent} or (ii) recognize degradations and then search the action space via exhaustive, greedy, or prior knowledge-based strategies \cite{agenticir, multiAgent, QAgent, hybridagent}.
As discussed in Sec.~\ref{section21}, these approaches incur substantial inference latency due to exhaustive tool searching and frequent rollbacks, and their performance depends critically on supervised degradation labels for recognition model training.
Our goal is to replace such search-heavy pipelines with a policy optimization-driven agent, named SimpeCall,  trained in a label-free environment using feedback from MLLMs.
The overall SimpeCall architecture is illustrated in Figure \ref{main4_fig}, where an restoration agent interacts with an environment that executes restoration actions, receives MLLM-based qualitative feedback, and learns an efficient restoration policy.

\subsection{Image Restoration Agent}
\label{ira_}
Our Restore-R1 agent performs two tasks: network update and action prediction.
Let a restoration trajectory be:
\begin{equation}
\zeta:=\{s_0, f_0, s_1, f_1, \ldots, s_{T}\},
\end{equation}
where $s_0=x_L$, $s_{t+1}=f_t(s_t)$, $s_t$ and $T$ denotes the maximum allowable length of the restoration sequence.
Let $\pi_{\theta}(f|s)$ denote a policy (actor) network parameterized by $\theta$, which governs the agent’s behavior by predicting a state-conditioned action distribution. Our objective is to optimize the expected return $R(\zeta)$:
\begin{equation}
\max_\theta \mathcal{L}(\theta)=  \mathbb{E}_{\zeta \sim \pi_\theta} [R(\zeta)] =\sum_\zeta P(\zeta \mid \theta) \ R(\zeta),
\end{equation}
where $R(\zeta):=\sum_{t=0}^T R(s_t, f_t)$ is the cumulative rewards, $R(s_t, f_t)$ denotes the immediate reward associated with executing action $f_t$ in state $s_t$, and $P(\zeta\ |\ \theta)$ is the probability of experiencing trajectory $\zeta$ under the policy $\pi_{\theta}$.

\textbf{Network Update}.
To optimize the policy network $\pi_\theta$, we let the agent interact with the environment to collect trajectories, receive reward signals from an evaluator (see Sec. \ref{LF_ENV}), and update the network parameters through policy optimization. 
A naïve policy objective function would be \cite{PPO}:
\begin{equation}
\mathcal{L}^{}(\theta) = \mathbb{E}_{t} \left[ \log \pi_\theta(f_t \mid s_t) \hat{A}_t \right],
\end{equation}
where $\pi_\theta(f_t \mid s_t)$ denotes the likelihood that the policy selects action $f_t$ given state $s_t$, and $\hat{A}_t$ is the corresponding advantage, where $\hat{A}_t>0$ means the action is better than other action possible at that state. 
To stabilize learning, we adopt the clipped surrogate objective function \cite{PPO}:
\begin{equation}
\mathcal{L}_{\text{PO}}(\theta) = \mathbb{E}_{t} \left[ \min \left( r_t(\theta) \hat{A}_t, \text{clip}(r_t(\theta), 1-\epsilon, 1+\epsilon) \hat{A}_t \right) \right], \nonumber
\end{equation}
where the scalar $\epsilon$ limits the magnitude of policy updates, and $r_t(\theta) = \frac{\pi_\theta(f_t \mid s_t)}{\pi_{\theta_{\text{old}}}(f_t \mid s_t)}$ is the importance sampling ratio, which measures how the action probability at state $s_t$ changes from the old policy to the current one.
A positive value of $r_t(\theta)$ indicates that action $f_t$ is favored more strongly by the current policy compared to the earlier one.
Clipping the ratio prevents overly large policy deviations, enforcing a soft trust region analogous to TRPO \cite{TRPO} while remaining computationally simple.

To evaluate the quality of the chosen action, we follow \cite{TRPO, PPO, GRPO} and maintain a critic network $V_\phi$ trained via:
\begin{equation}
\mathcal{L}_{\text{VF}}(\phi) = \mathbb{E}_{t} \left[ \left( V_\phi(s_t) - \hat{R}_t \right)^2 \right],
\end{equation}
where $\hat{R}_t$ is an estimate of the return. 
We then compute $\hat{A}_t$ using generalized advantage estimation (GAE) \cite{GAE}:
\begin{equation}
\hat{A}_t = \sum_{l=0}^{\infty} (\gamma \lambda)^l \delta_{t+l},  \ \delta_t = R(s_t, f_t)+\gamma V(s_{t+1})-V(s_t), \nonumber
\end{equation}
where $\gamma$ discounts future rewards and $\lambda$ controls the GAE weighting.
To encourage exploration, we measure the uncertainty of the policy distribution $\mathcal{H}[\cdot]$ with an entropy bonus term:
\begin{equation}
\mathcal{L}_{\text{EB}}(\theta) = \mathbb{E}_{t} \left[ \mathcal{H} \left[ \pi_\theta(\cdot \mid s_t) \right] \right],
\end{equation}
Then the final objective can be written as:
\begin{equation}
\mathcal{L} =  \mathcal{L}_{\text{PO}} - c_1 \cdot \mathcal{L}_{\text{VF}} + c_2 \cdot \mathcal{L}_{\text{EB}},
\end{equation}
where $c_1$ and $c_2$ are balancing coefficients.

\textbf{Action Prediction}.
During training, actions are sampled from the stochastic policy, i.e., $f_t \sim \pi_\theta(\cdot | s_t)$, to encourage exploration and collect diverse trajectories.
In contrast, during inference, the agent adopts a deterministic strategy by selecting the most probable action, i.e., $f_t=\arg\max_f \pi_\theta(f|s_t)$.
The policy network comprises a vision encoder followed by a lightweight multilayer perceptron (MLP).
Each state $s_t$ is represented by concatenating the visual features extracted from the current input with an action record vector that summarizes the policy’s past decisions.
Given this state representation, the actor predicts the next restoration tool, and this process is repeated iteratively until the image converges to a visually satisfactory result.

\begin{table*}[!tb]
\centering
\small
\setlength{\tabcolsep}{3.5pt}
\caption{Comparison with SOTA methods on \underline{F}ull-reference metrics: {F1} (PSNR), {F2} (SSIM), {F3} (LPIPS), 
and \underline{N}o-reference metrics: {N1} (MANIQA), {N2} (CLIP-IQA), {N3} (MUSIQ), {N4} (DeQA-Score) across all settings. ``GT'' denotes whether ground truth supervision is required for model training. Remarkably, \textbf{without using any labels}, our method matches SOTA performance on full-reference metrics and outperforms compared methods on no-reference metrics. }
\begin{tabular}{l|c|ccccccc|ccccccc}
\toprule
\multirow{2}{*}{Method} &\multirow{2}{*}{GT}  & \multicolumn{3}{>{\columncolor{cyan!10}}c}{Full-Reference}  & \multicolumn{4}{>{\columncolor{green!10}}c|}{No-Reference}  & \multicolumn{3}{>{\columncolor{cyan!10}}c}{Full-Reference}  & \multicolumn{4}{>{\columncolor{green!10}}c}{No-Reference} \\
& & \cellcolor{cyan!10}{F1}~↑ & \cellcolor{cyan!10}{F2}~↑ & \cellcolor{cyan!10}{F3}~↓ & \cellcolor{green!10}{N1}~↑ & \cellcolor{green!10}{N2}~↑ & \cellcolor{green!10}{N3}~↑ & \cellcolor{green!10}{N4}~↑ & \cellcolor{cyan!10}{F1}~↑ & \cellcolor{cyan!10}{F2}~↑ & \cellcolor{cyan!10}{F3}~↓ & \cellcolor{green!10}{N1}~↑ & \cellcolor{green!10}{N2}~↑ & \cellcolor{green!10}{N3}~↑ & \cellcolor{green!10}{N4}~↑ \\
\midrule
& & \multicolumn{7}{c|}{Setting I}
& \multicolumn{7}{c}{Setting II }  \\
AirNet \cite{airnet} &\ding{51}  &18.489	&0.608	&0.374	&0.295	&0.399	&48.850	&2.911 &17.196	&0.642	&0.326	&0.323	&0.439	&53.794	&3.098 \\
PromptIR \cite{Promptir} &\ding{51}  &19.198	&0.609	&0.371	&0.300	&0.400	&49.189	&2.907 &17.485	&0.660	&0.301	&\cellcolor{green!10}0.324	&0.441	&53.823	&3.154 \\
InstructIR \cite{Instructir} &\ding{51}  &18.394	&0.586	&0.416	&0.268	&0.390	&46.047	&2.844 &17.508	&0.593	&0.412	&0.286	&0.433	&48.147	&2.994 \\
MiOIR(R) \cite{kong2024towards} &\ding{51}  &\cellcolor{cyan!10}19.970	&0.659	&0.337	&0.282	&0.430	&50.269	&2.998 &18.045	&0.671	&0.304	&0.288	&\cellcolor{green!10}0.479	&53.303	&3.173  \\
MiOIR(U) \cite{kong2024towards} &\ding{51}  &19.920	&\cellcolor{cyan!10}0.672	&\cellcolor{cyan!10}0.332	&0.289	&0.436	&51.217	&3.011 &17.786	&\cellcolor{cyan!10}0.676	&\cellcolor{cyan!25}0.278	&0.299	&0.466	&55.143	&3.191 
 \\
DA-CLIP \cite{daclip} &\ding{51}  &19.252	&0.614	&0.370	&0.292	&0.417	&50.320	&2.974 &17.695	&0.637	&0.349	&0.308	&0.414	&53.729	&3.071 \\
AutoDIR \cite{Autodir} &\ding{51}  &19.405	&0.644	&0.346	&0.306	&0.431	&53.580	&3.186 &\cellcolor{cyan!10}18.540	&0.664	&0.319	&0.318	&0.448	&55.683	&3.257 \\
AgenticIR \cite{agenticir} &\ding{51}   & \cellcolor{cyan!25}20.923 & \cellcolor{cyan!25}0.698 & \cellcolor{cyan!25}0.306 & \cellcolor{green!10}0.315 & \cellcolor{green!10}0.441 & \cellcolor{green!10}58.590 & \cellcolor{green!10}3.423 & \cellcolor{cyan!25}20.418 & \cellcolor{cyan!25}0.719 & \cellcolor{cyan!10}0.300 & 0.310 & 0.448 & \cellcolor{green!10}58.643 & \cellcolor{green!10}3.434  \\
Ours &\ding{55}  & 19.513 & 0.647 & 0.365 & \cellcolor{green!25}0.349 & \cellcolor{green!25}0.525 & \cellcolor{green!25}63.003 & \cellcolor{green!25}3.657 & 17.958 & 0.636 & 0.385 & \cellcolor{green!25}0.343 & \cellcolor{green!25}0.530 & \cellcolor{green!25}63.344 & \cellcolor{green!25}3.599  \\
\midrule
& & \multicolumn{7}{c|}{Setting III} & \multicolumn{7}{c}{Setting IV}  \\
AirNet \cite{airnet} &\ding{51}  &16.469	&0.517	&0.504	&\cellcolor{green!10}0.271	&0.349	&45.650	&2.540  &16.128	&0.472	&0.606	&0.155	&0.197	&24.761	&1.941  \\
PromptIR \cite{Promptir} &\ding{51}  &16.513	&0.511	&0.515	&0.262	&0.343	&44.741	&2.536 &\cellcolor{cyan!10}16.692	&\cellcolor{cyan!10}0.492	&0.594	&0.168	&0.212	&24.830 &1.987 \\
InstructIR \cite{Instructir} &\ding{51}  &15.912	&0.414	&0.664	&0.192	&0.343	&34.712	&2.331 &15.702	&0.488	&0.632	&0.178	&0.192	&25.373	&2.008 
 \\
MiOIR(R) \cite{kong2024towards} &\ding{51}  &16.575	&0.514	&0.512	&0.215	&0.394	&41.312	&2.608 &15.608	&0.487	&0.633	&0.178	&0.225	&26.632	&2.030 
 \\
MiOIR(U) \cite{kong2024towards} &\ding{51}  &16.599	&0.529	&0.505	&0.237	&0.376	&43.873	&2.586 &15.596	&0.488	&0.643	&0.174	&0.254	&26.902	&2.022 
 \\
DA-CLIP \cite{daclip} &\ding{51}  &16.388	&0.470	&0.562	&0.233	&0.342	&41.088	&2.488 &15.491	&0.482	&0.626	&0.181	&0.236	&26.862	&2.048 
 \\
AutoDIR \cite{Autodir} &\ding{51}  &16.870	&0.540	&\cellcolor{cyan!25}0.451	&0.263	&\cellcolor{green!10}0.396	&\cellcolor{green!10}49.707	&\cellcolor{green!10}2.867 &15.651	&0.464	&0.582	&\cellcolor{green!10}0.203	&0.231	&32.172	&2.199  \\
AgenticIR \cite{agenticir} &\ding{51}   & \cellcolor{cyan!25}18.600 & \cellcolor{cyan!25}0.601 & \cellcolor{cyan!10}0.465 & 0.235 & 0.337 & 47.811 & 2.836  & \cellcolor{cyan!25}16.879 & \cellcolor{cyan!25}0.498 & \cellcolor{cyan!10}0.573 & 0.165 & \cellcolor{green!10}0.259 & \cellcolor{green!10}37.100 & \cellcolor{green!10}2.207 \\
Ours &\ding{55}  & \cellcolor{cyan!10}17.650 & \cellcolor{cyan!10}0.563 & 0.475 & \cellcolor{green!25}0.295 & \cellcolor{green!25}0.473 & \cellcolor{green!25}57.756 & \cellcolor{green!25}3.247  & 16.180 & 0.473 & \cellcolor{cyan!25}0.564 & \cellcolor{green!25}0.233 & \cellcolor{green!25}0.389 & \cellcolor{green!25}49.810 & \cellcolor{green!25}2.864  \\
\bottomrule
\end{tabular}
\label{sota_comparison}
\end{table*}

\subsection{Label-Free Environment}
\label{LF_ENV}
The environment is responsible for action execution and evaluation.
It applies a selected restoration tool to the input image, and then evaluates the output to provide feedback signal for reward computation.

\textbf{Action Evaluation}.
Existing restoration agents \cite{RestoreAgent, agenticir, multiAgent, QAgent, hybridagent} adopt Llava-Llama3 \cite{touvron2023llama}, DepictQA \cite{DepictQA}, Co-instruct \cite{wu2024towards}, and Qwen2-VL \cite{wang2024qwen2} to recognize degradation types and assess restoration quality.
However, these models requires extensive labeled data for supervised training.
In our label-free environment, neither clean images nor ground-truth optimal tool sequences are available, which renders supervised quality prediction infeasible.
To overcome this challenge, we leverage the perceptual capability of MLLMs \cite{dong2026refadv, lu2025representation, you2025teaching, DepictQA} as a source of reward feedback. 
A key difficulty is that most MLLMs generate discrete textual tokens, which do not align with the continuous nature of restoration quality and thus are unsuitable for reward computation. 
Rather than forcing MLLMs to output discretized scores, we seek a model capable of producing a continuous quality distribution that can serve as numerical feedback for policy optimization.
To achieve this, we adopt the recently proposed DeQA-Score \cite{you2025teaching} as the evaluator, which produces distributional quality estimates compatible with reward shaping.
During training, the reward at step $t$ is defined as the improvement in DeQA-Score:
\begin{equation}
R(s_t, f_t) = \text{DS}(s_{t+1}) - \text{DS}(s_t),
\end{equation}
where $\text{DS}(\cdot)$ denotes the DeQA-Score evaluator. 
Positive values indicate that the selected tool improves perceptual quality, while negative values penalize harmful or unnecessary actions.
This formulation converts MLLM perception into a practical and stable reward signal, enabling fully label-free policy learning.

\begin{figure*}
\centering
\includegraphics[scale=0.69,trim=2 0 2 0,clip]{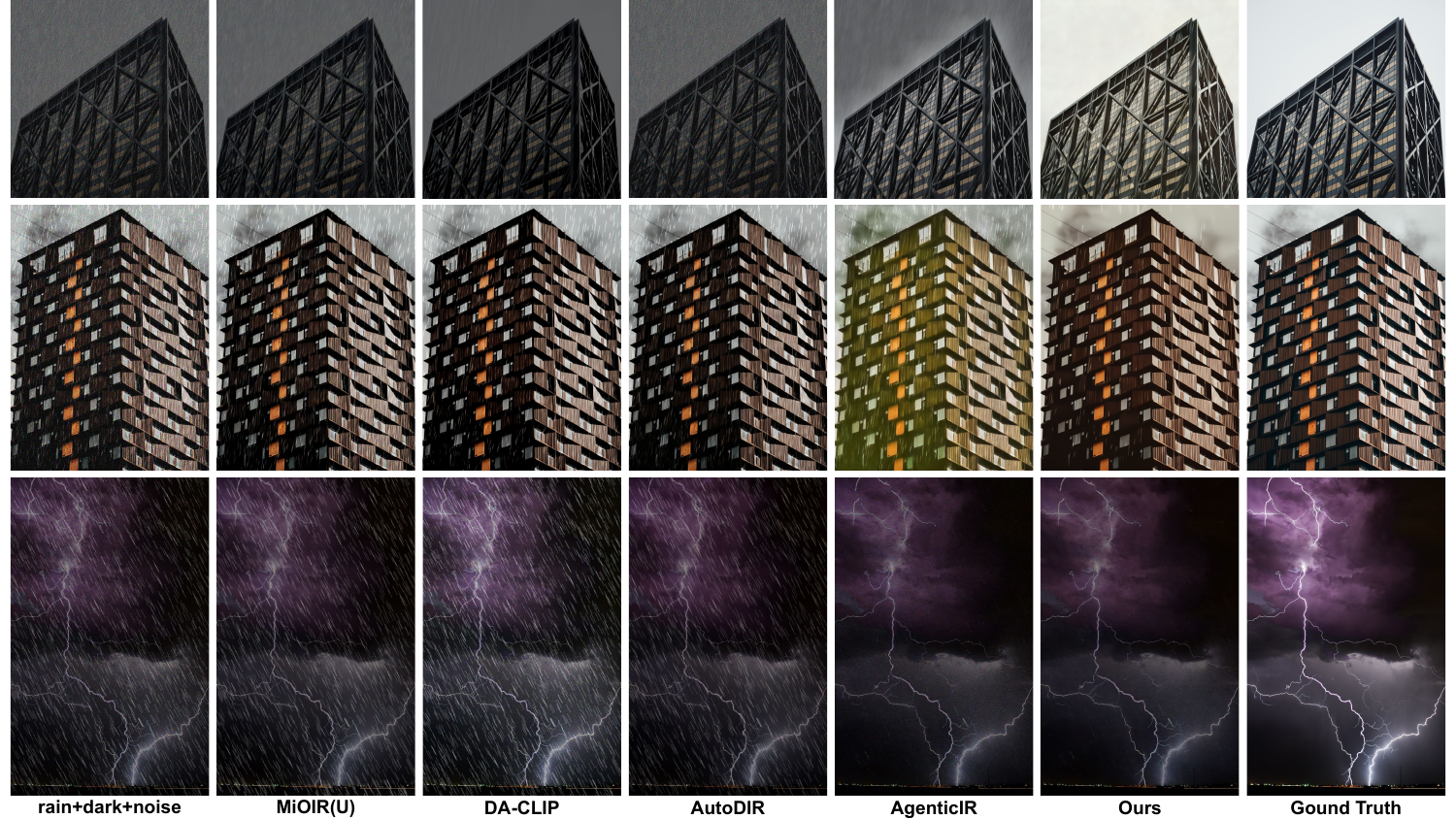}
\includegraphics[scale=0.69,trim=2 0 2 150,clip]{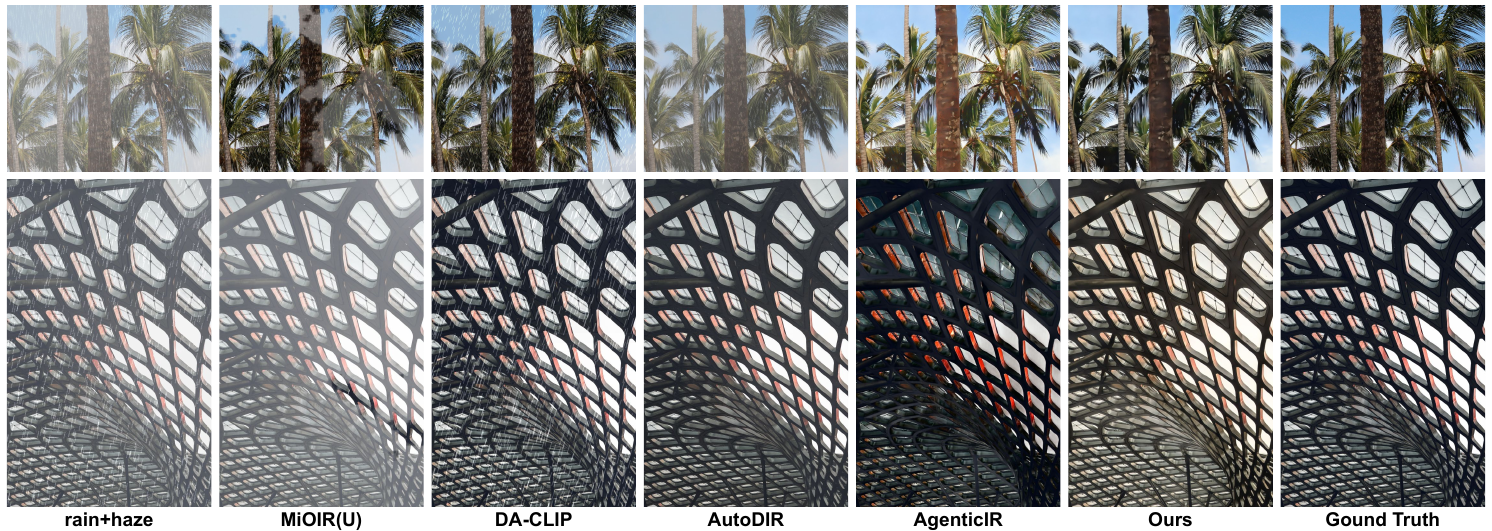}
\caption{Qualitative comparison between our method and SOTA restoration baselines (for other baselines see the supplementary material).}
\label{vis_compare111}
\end{figure*}

\textbf{Action Execution}.
We consider seven degradation types commonly found in real-world CIR, including haze, defocus blur, motion blur, rain, noise, dark, and JPEG compression artifact. For these degradations, we construct a tool set for the agent to use, which consist of multiple specialized models 
for each type of degradation. 
Following \cite{agenticir, multiAgent}, we collect three to six publicly available restoration models for each degradation. 
For example, 
we take MAXIM \cite{Maxim},
X-Restormer \cite{Xrestormer};
RIDCP \cite{RIDCP},
DehazeFormer \cite{DehazeFormer} as the available tools for Dehazing. 
A detailed description of the tool set can be seen in the Supplementary Material.
This tool set forms the full discrete action space for the policy.

%% file: sec/4_experiments.tex
\section{Experiments}
\label{experiments}

\subsection{Experimental Settings}
\textbf{Data Construction}.
To train a restoration agent capable of handling diverse degradations, we construct a dataset that includes mixed-degradation images with varying complexity.
Specifically, we design 15 degradation combinations, grouped into four settings, including Setting I (5 combinations of 2 degradations), Setting II (3 combinations of 2 degradations), Setting III (4 combinations of 3 degradations), and Setting IV (3 combinations of 4 or 5 degradations). 
Details of all combinations are provided in the supplementary material. 
Following previous agentic studies 
\cite{agenticir, multiAgent}, we use clean images from the MiO100 dataset \cite{kong2024towards, kong2024preliminary} to generate degraded images.
For training, we only use images from Setting I, sampling 20 images per combination to construct the training set, while all remaining images, including all images from Setting II-IV, are reserved for testing.
This ensures that out agent is evaluated on unseen degradation mixtures, enabling a rigorous assessment of its generalization capability.

\begin{table}[!tb]
\centering
\small
\setlength{\tabcolsep}{2.1pt}
\caption{Performance comparison across different reward strategies. ``Ours (N$_i$)" denotes the use of the $i$-th no-reference quality assessment model for reward computation. Results are reported on \underline{F}ull-reference metrics: {F1} (PSNR), {F2} (SSIM), {F3} (LPIPS), and \underline{N}o-reference metrics: {N1} (MANIQA), {N2} (CLIP-IQA), {N3} (MUSIQ), and {N4} (DeQA-Score).}
\begin{tabular}{c|l|ccccccc}
\toprule
& \multirow{2}{*}{Rewards}  & \multicolumn{3}{>{\columncolor{cyan!10}}c}{Full-Reference}  & \multicolumn{4}{>{\columncolor{green!10}}c}{No-Reference} \\
& & \cellcolor{cyan!10}{F1}~↑ & \cellcolor{cyan!10}{F2}~↑ & \cellcolor{cyan!10}{F3}~↓ & \cellcolor{green!10}{N1}~↑ & \cellcolor{green!10}{N2}~↑ & \cellcolor{green!10}{N3}~↑ & \cellcolor{green!10}{N4}~↑ \\
\midrule
\multirow{4}{*}{\rotatebox{90}{Setting I}} 
& Ours ({N1})    
&\cellcolor{cyan!10}18.955	&0.602	&0.448	&\cellcolor{green!25}0.383	&0.476	&61.652	&3.316 \\
& Ours ({N2}) 
&18.340	&\cellcolor{cyan!10}0.610	&\cellcolor{cyan!10}0.411	&0.332	&\cellcolor{green!10}0.523	&60.129	&3.230 \\
& Ours ({N3}) 
&17.970	&0.602	&0.415	&\cellcolor{green!10}0.370	&0.503	&\cellcolor{green!10}62.754	&\cellcolor{green!10}3.380 \\
& Ours ({N4}) 
& \cellcolor{cyan!25}19.513 & \cellcolor{cyan!25}0.647 & \cellcolor{cyan!25}0.365 & 0.349 & \cellcolor{green!25}0.525 & \cellcolor{green!25}63.003 & \cellcolor{green!25}3.657 \\
\midrule
\multirow{4}{*}{\rotatebox{90}{Setting II}} 
& Ours ({N1}) &\cellcolor{cyan!10}17.364	&0.578	&0.456	&\cellcolor{green!25}0.371	&0.464	&61.578	&3.234 \\
& Ours ({N2}) &16.991	&0.580	&0.432	&0.332	&\cellcolor{green!25}0.536	&\cellcolor{green!10}62.300	&\cellcolor{green!10}3.252 \\
& Ours ({N3}) &16.762	&\cellcolor{cyan!10}0.584	&\cellcolor{cyan!10}0.422	&0.331	&0.466	&60.711	&3.220 \\
& Ours ({N4})  & \cellcolor{cyan!25}17.958 & \cellcolor{cyan!25}0.636 & \cellcolor{cyan!25}0.385 & \cellcolor{green!10}0.343 & \cellcolor{green!10}0.530 & \cellcolor{green!25}63.344 & \cellcolor{green!25}3.599 \\
\midrule
\multirow{4}{*}{\rotatebox{90}{Setting III}} 
& Ours ({N1}) &\cellcolor{cyan!10}17.418	&\cellcolor{cyan!10}0.538	&\cellcolor{cyan!10}0.517	&\cellcolor{green!25}0.343	&0.444	&\cellcolor{green!25}58.988	&\cellcolor{green!10}3.048 \\
& Ours ({N2}) &16.571	&0.462	&0.549	&\cellcolor{green!10}0.328	&\cellcolor{green!25}0.570	&57.727	&2.824 \\
& Ours ({N3}) &16.989	&0.515	&0.526	&0.311	&0.460	&55.076	&2.857 \\
& Ours ({N4}) & \cellcolor{cyan!25}17.650 & \cellcolor{cyan!25}0.563 & \cellcolor{cyan!25}0.475 & 0.295 & \cellcolor{green!10}0.473 & \cellcolor{green!10}57.756 & \cellcolor{green!25}3.247 \\
\midrule
\multirow{4}{*}{\rotatebox{90}{Setting IV}} 
& Ours ({N1}) &\cellcolor{cyan!10}15.813	&\cellcolor{cyan!25}0.475	&0.599	&\cellcolor{green!25}0.271	&0.342	&\cellcolor{green!25}49.895	&\cellcolor{green!10}2.592 \\
& Ours ({N2}) &15.091	&0.445	&\cellcolor{cyan!25}0.563	&0.213	&\cellcolor{green!10}0.371	&48.036	&2.471 \\
& Ours ({N3}) &15.449	&0.455	&0.570	&0.227	&0.261	&44.688	&2.279 \\
& Ours ({N4})  & \cellcolor{cyan!25}16.180 & \cellcolor{cyan!10}0.473 & \cellcolor{cyan!10}0.564 & \cellcolor{green!10}0.233 & \cellcolor{green!25}0.389 & \cellcolor{green!10}49.810 & \cellcolor{green!25}2.864 \\
\bottomrule
\end{tabular}
\label{rewards}
\end{table}

\textbf{Baselines}.
We compare our method against both all-in-one and agentic approaches. All-in-one methods include AirNet \cite{airnet}, PromptIR \cite{Promptir}, InstructIR \cite{Instructir}, MiOIR(R) \cite{kong2024towards} (with Restormer \cite{Restormer}), MiOIR(U) \cite{kong2024towards} (with Uformer \cite{Uformer}), DA-CLIP \cite{daclip}, and AutoDIR \cite{Autodir}. 
For agentic baselines, we consider the SOTA method AgenticIR \cite{agenticir}.
Other agentic approaches \cite{RestoreAgent, multiAgent, QAgent, hybridagent} have not released code or checkpoints, but from our analysis in Sec. \ref{section21} and summarization in Table \ref{agents_comparison}, they share similar workflow and are expected to perform comparably to AgenticIR.
Notably, all these methods require ground-truths for model training.

\textbf{Implementation Details}.
For model construction, the actor and critic share a frozen CLIP (ViT-L/14) vision encoder \cite{radford2021learning}, while each uses its own two-layer MLP with LayerNorm and ReLU.
The actor predicts a probability distribution across the set of available actions, whereas the critic produces a scalar value estimate. Both networks use a hidden dimension of 128.
The state representation is formed by concatenating visual features with a binary action-record vector derived from the previous step’s action distribution.
For model training, we set the learning rate, critic coefficient, entropy coefficient, entropy decay factor, discount factor, GAE parameter, and clipping factor to 0.01, 0.5, 0.05, 0.99, 0.99, 0.95, and 0.2, respectively.
The complete restoration tool set is described in supplementary material.

\textbf{Evaluation Metrics}.
We evaluate models using seven quality metrics, including three full-reference metrics: PSNR, SSIM \cite{SSIM}, LPIPS \cite{LPIPS}, and four no-reference metrics: MANIQA \cite{MANIQA}, CLIP-IQA \cite{CLIPIQA}, MUSIQ \cite{MUSIQ}, and DeQA-Score \cite{you2025teaching}. 
Detailed definitions and properties of these metrics are provided in the supplementary material.

\subsection{Comparison with SOTA}
\textbf{Quantitative Comparison}.
Table \ref{sota_comparison} presents the performance comparison between our method and compared baselines across all degradation settings. 
Despite not relying no any ground-truth supervision, our method achieves highly competitive performance on full-reference metrics across all degradation settings.
This demonstrates the ability of our agent to perform faithful structural restoration guided only by perception feedback.
More notably, as degradation complexity increases (Settings III and IV), our method demonstrates superior robustness and consistently surpasses all baselines on no-reference metrics, achieving the highest perceptual quality.
This suggests that our agent effectively learns restoration strategies that generalize well to unseen mixtures and emphasize perceptual coherence.

\begin{figure}
\centering
\includegraphics[scale=0.50,trim=10 10 16 25,clip]{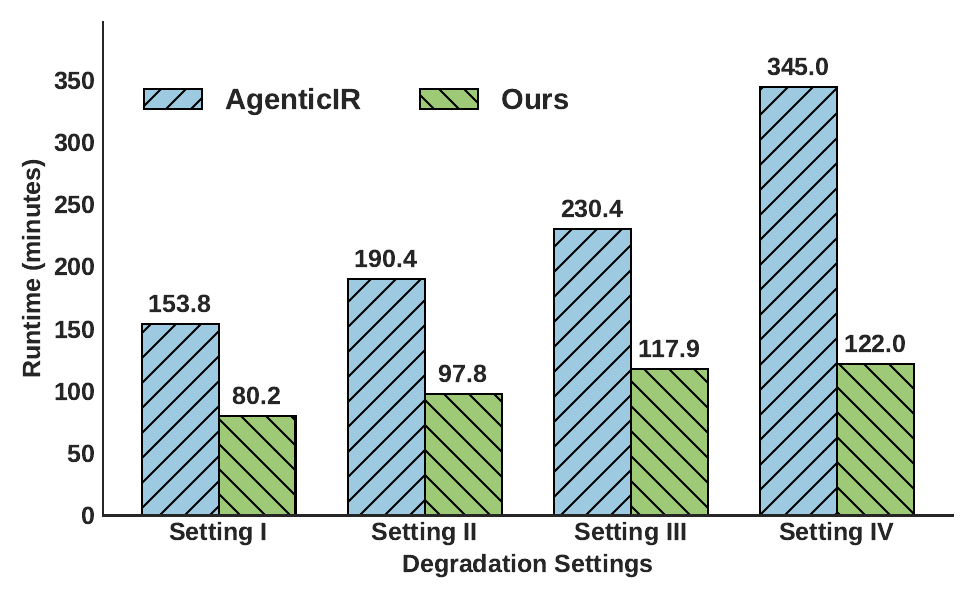}
\caption{Runtime comparison between ours and AgenticIR \cite{agenticir}.}
\label{fig_runtime}
\end{figure}

\begin{figure}
\centering
\includegraphics[scale=0.472,trim=110 100 110 100,clip]{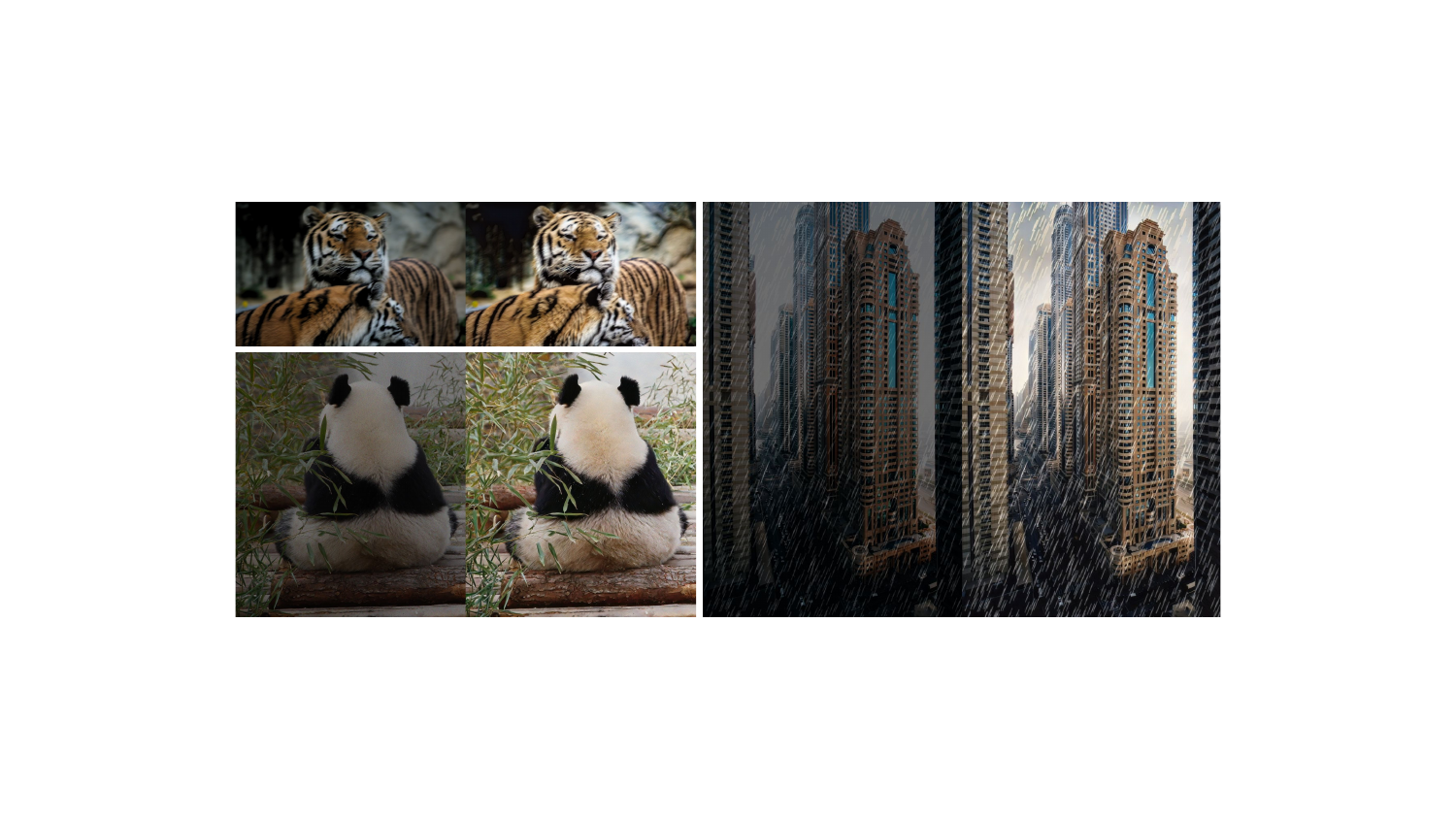}
\caption{Illustration of tool effects. Left: images with dark degradation (tiger: motion blur+dark, panda: dark+noise; skyscraper: rain+dark). Right: outputs from the dehazing model RIDCP \cite{RIDCP}.}
\label{tooleffect2}
\end{figure}

\begin{figure*}[!tb]
\centering
\begin{subfigure}{0.335\linewidth}
\includegraphics[scale=0.325,trim=7 8 8 6,clip]{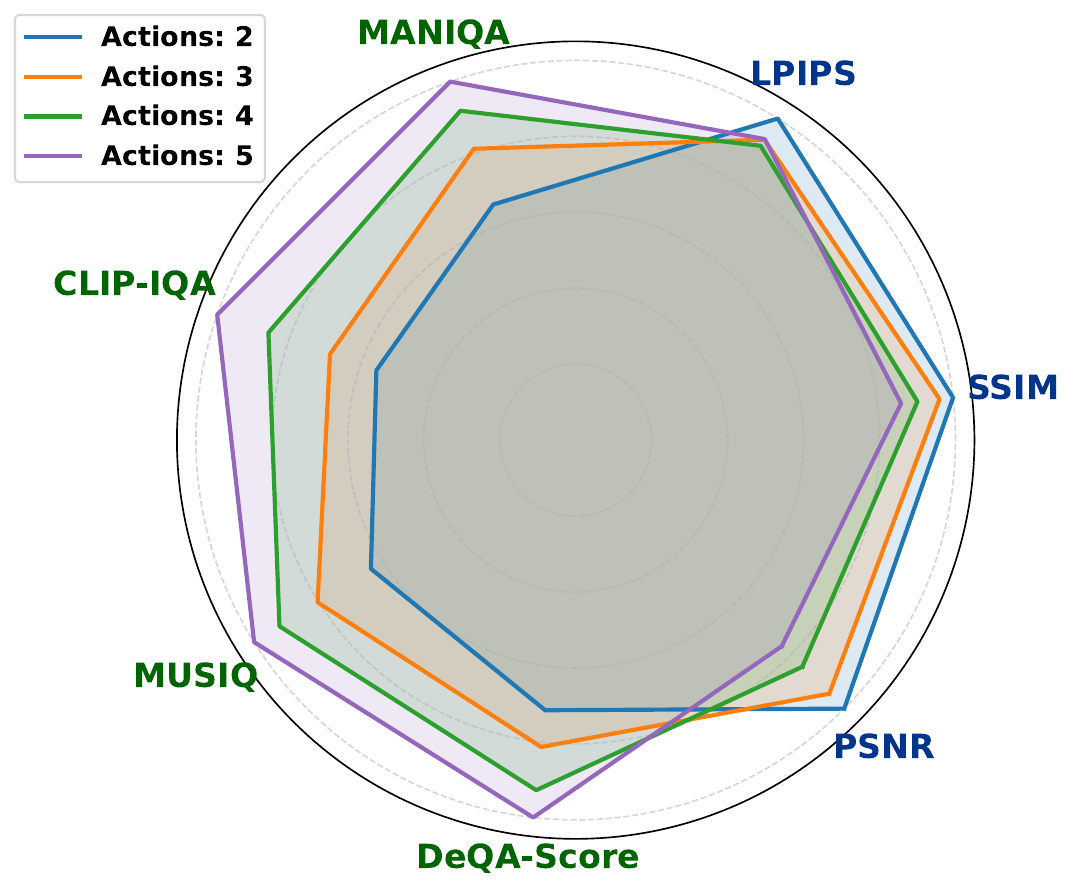}
\caption{3 degradations}
\label{subfig_a}
\end{subfigure}
\begin{subfigure}{0.315\linewidth}
\includegraphics[scale=0.325,trim=8 8 8 6,clip]{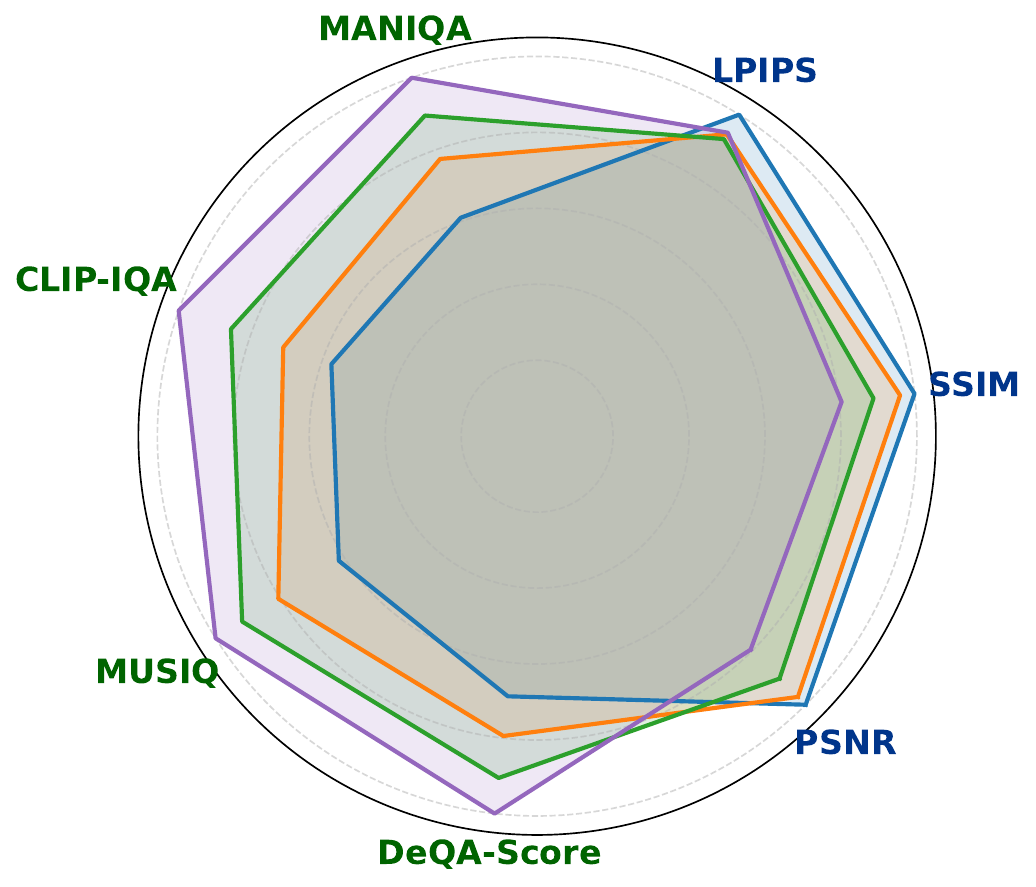}
\caption{5 degradations}
\label{subfig_b}
\end{subfigure}
\begin{subfigure}{0.335\linewidth}
\includegraphics[scale=0.325,trim=7 8 8 6,clip]{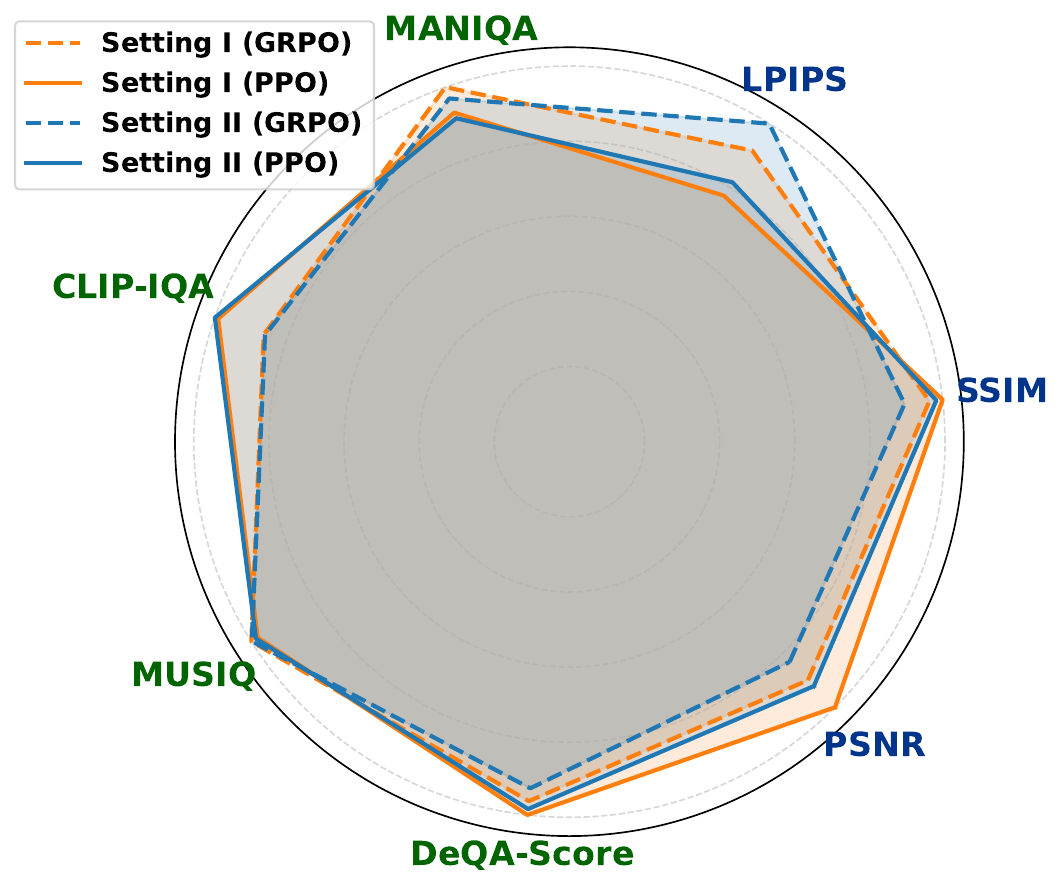}
\caption{GRPO vs. PPO}
\label{subfig_c}
\end{subfigure}
\caption{Illustration of the distortion-perception tradeoff for (a) 3 degradations and (b) 5 degradations. As the number of actions increases, the circles in the radar plots gradually shift leftward. (c) Performance comparison between PPO and GRPO on two different settings.}
\label{fig_s1}
\end{figure*}

\textbf{Qualitative Comparison}.
Figure \ref{vis_compare111} presents visual comparisons under rain+haze and rain+dark+noise degradations. The comparison with other baselines are provided in the supplementary. 
Despite the absence of degradation annotations, our method learns to remove multiple corruptions from degraded images and achieves comparable or superior visual quality than the supervised baselines. This shows the effectiveness of our proposed MLLM perceptual feedback.

\textbf{Efficiency Comparison}.
Figure \ref{fig_runtime} presents the runtime comparison between our method and AgenticIR \cite{agenticir} under all settings.
As seen, AgenticIR’s inference time grows rapidly from Setting I (2 degradations) to Setting IV (4 or 5 degradations),
This trend is expected because AgenticIR conducts tool searching for each degradation type, causing runtime to scale with the number of degradations present in the input. 
Moreover, its reliance on reflection and rollback introduces additional tool invocations, further compounding the overall computational overhead.
In contrast, our inference time remains almost constant across all settings.
The efficiency originates from our one-pass design: the agent outputs a complete tool-calling sequence in a single policy execution, without any iterative search or backtracking.

\subsection{Ablation Studies}
\label{AblationStudies}

\textbf{Selection on Reward Functions}.
Different restoration tools contribute unevenly depending on the evaluation perspective, which directly influences reward computation and policy learning.
To investigate this effect, we present ablation results under various reward configurations in Table \ref{rewards}.
While MANIQA and CLIP-IQA rewards offer strong learning signals and yield competitive results, they show preference toward specific scenarios and lead to performance imbalance.
For instance, MANIQA performs well on Settings III-IV but poorly on Settings I-II. 
In contrast, DeQA-Score rewards consistently yield the best or near-best results under full-reference and no-reference metrics.

\textbf{Effect of Tools}.
Prior agents \cite{RestoreAgent, agenticir, multiAgent} rely heavily on degradation recognition and strictly map each degradation type to a specific group of tools. 
However, they largely ignore the effects of tool interactions and degradation stacking. 
In practice, tools designed for a particular degradation type may also produce unexpected enhancement effects on other corruption types.
As shown in Figure~\ref{tooleffect2}, the dehazing model RIDCP~\cite{RIDCP}, although originally designed for haze removal, significantly improves visibility and brightness in low-light images. 
This observation indicates that the functional scope of a tool may extend beyond its intended degradation domain, due to inherent model capacity, interactions of tools, or degradation stacking effects.
Therefore, the optimal tool sequence might not be inferred from a simple type-to-tool mapping.
This motivates the need for our policy-based approach that learns tool sequencing holistically rather than relying on predefined associations.

\textbf{Perception-Distortion Tradeoff}.
Figures \ref{subfig_a} and \ref{subfig_b} show that as the number of execution steps increases, the no-reference metrics continue to improve, whereas the full-reference metrics gradually decline (additional results are provided in the supplementary material).
The rise in no-reference scores is expected because our agent is optimized to maximize the accumulated final-stage rewards, which is tied to perceptual quality.
The concurrent drop in full-reference metrics aligns with observations in prior work \cite{blau2018perception}, which established the theory, termed the perception-distortion tradeoff.
This theory demonstrates that no image restoration algorithm can simultaneously achieve both minimal distortion and the maximal perceptual quality; improving one inevitably degrades the other due to the intrinsic ambiguity of the inverse imaging process.

\textbf{GRPO Optimization}.
Our framework is compatible with  various policy optimization methods. To demonstrate this flexibility, we additionally train the model using GRPO \cite{GRPO}. Figure \ref{subfig_c} compares the results of GRPO and PPO, showing that both training strategies lead to similar performance. 
This confirms that our framework generalizes well across different policy optimization algorithms.
Due to space limitation, we provide additional empirical results in the supplementary material.

%% file: sec/5_conclusion.tex
\section{Conclusion}
In this work, we propose a policy optimization-based framework for complex image restoration, in which a lightweight agent learns to select restoration tools and determine their execution order through interaction with multimodal perceptual feedback.  
Compared with existing restoration approaches, our method eliminates the reliance on reflection, rollback, and tool searching, improving inference efficiency while maintaining high restoration quality. 
We believe this work represents a promising step toward autonomous, perception-driven restoration systems that can adapt to general conditions without explicit supervision. 

%% file: sec/X_suppl.tex
\clearpage
\setcounter{page}{1}
\maketitlesupplementary

\begin{table*}[!tb]
\centering
\small
\setlength{\tabcolsep}{8.0pt}
\caption{Degradation data construction}
\begin{tabular}{ccll}
\toprule
Settings  & \# of Degradations & Case Number & Combinations \\
\midrule
\multirow{5}{*}{I} & \multirow{5}{*}{2} & Case 1 & dark+noise \\
& & Case 2 & defocus blur+JPEG compression artifact \\
& & Case 3 & motion blur + dark \\
& & Case 4 & noise+JPEG compression artifact \\
& & Case 5 & rain+haze \\
\midrule
\multirow{3}{*}{II} & \multirow{3}{*}{2} & Case 6 & haze+noise \\
& & Case 7 & motion blur+JPEG compression artifact \\
& & Case 8 & rain+dark \\
\midrule
\multirow{4}{*}{III} & \multirow{4}{*}{3}  & Case 9 & dark+defocus blur+JPEG compression artifact
\\
& & Case 10 & motion blur+defocus blur+noise \\
& & Case 11 & rain+dark+noise \\
& & Case 12 & rain+haze+noise \\
\midrule
\multirow{3}{*}{IV} & \multirow{3}{*}{$>$3} & Case 13 & haze+dark+motion blur+JPEG compression artifact \\
& & Case 14 & rain+haze+defocus blur+JPEG compression artifact \\
& & Case 15 & rain+motion blur+defocus blur+noise+JPEG compression artifact \\
\bottomrule
\end{tabular}
\label{tab_degradations}
\end{table*}

\begin{figure*}[!tb]
\centering
\begin{subfigure}{0.48\linewidth}
\includegraphics[scale=0.43,trim=6 8 9 6,clip]{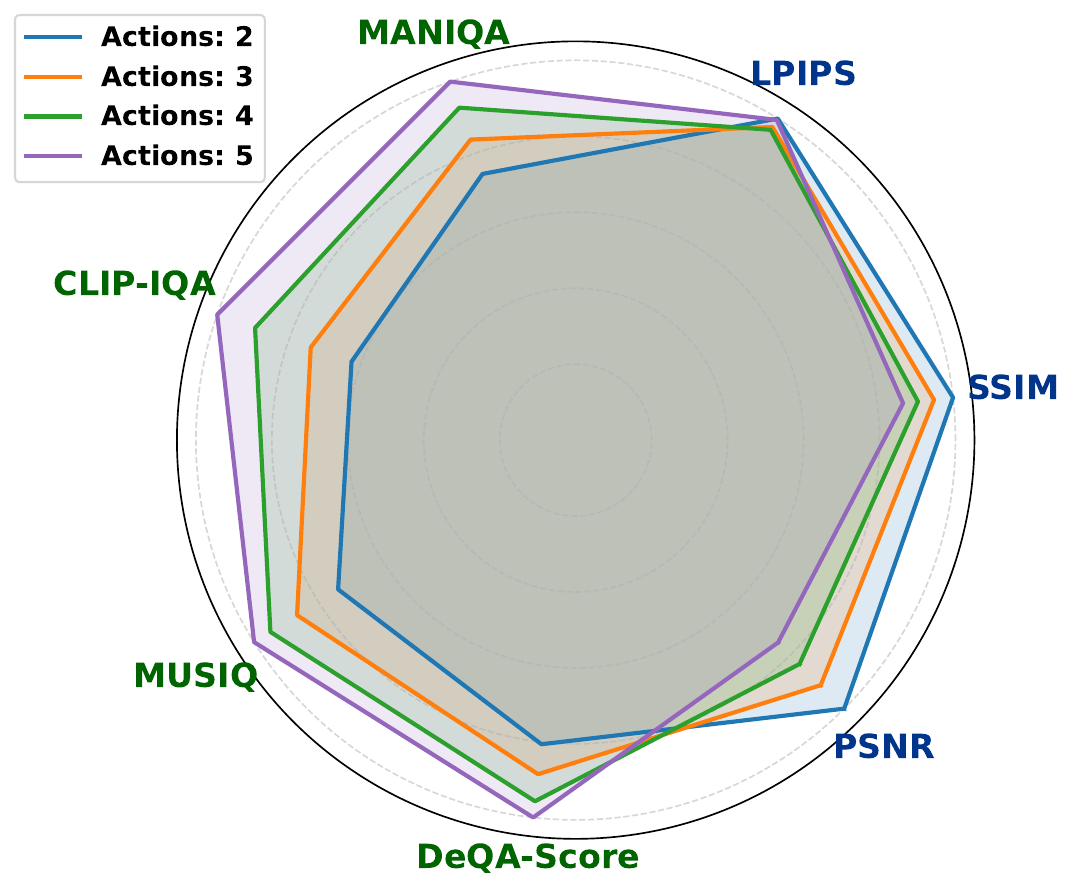}
\caption{defocus blur+JPEG compression artifact}
\label{supp_trade_off1}
\end{subfigure}
\begin{subfigure}{0.48\linewidth}
\includegraphics[scale=0.43,trim=9 8 8 6,clip]{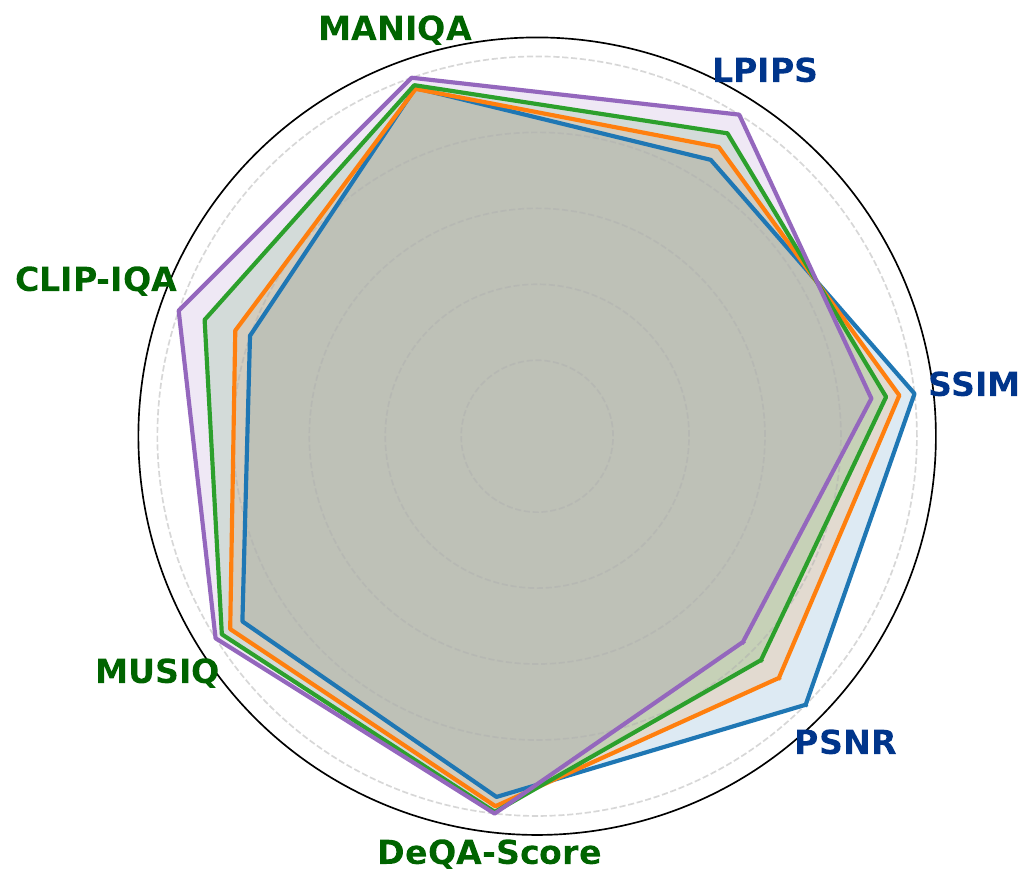}
\caption{noise+JPEG compression artifact} 
\label{supp_trade_off2}
\end{subfigure}
\caption{Illustration of distortion-perception tradeoff on (a) noise+jpeg compression artifact and (b) motion blur+defocus blur+noise.}
\end{figure*}

\section{Experimental Details}
\subsection{Data}
\label{Supply_Degradation}
In this section, we show how to synthesize degraded images following existing work \cite{agenticir}.
For dark images, the V channel value of the images in the HSV color space will be randomly decreased by one of the following strategies: linear mapping, Gamma correction, and subtracting a constant.
For defocus blur, the images will be filtered with circular kernels with random radius as in \cite{michaelis2019benchmarking}. 
For JPEG compression artifact, images are compressed with random quality factor, such as 5, 40, and 90. 
For noise, images are added with Poisson or Gaussian noise with random scale.
For rain, images are first added with noise and filtered the noise with linear kernels with random directions as in \cite{kong2024towards}. 
For motion blur, images are filtered with linear kernels with random direction and radius as in \cite{michaelis2019benchmarking}.
For haze generation, we simulate degraded images using the atmospheric scattering model, where the global atmospheric light and scattering coefficient are randomly sampled following the settings in prior work \cite{he2010single, li2018benchmarking}.
Following \cite{agenticir}, we consider degradation combinations that are common in real-world scenarios, e.g., dark+noise, rain+haze, rain+dark, rain+haze+defocus blur + JPEG compression artifact.
Table \ref{tab_degradations} shows the 15 degradation combinations used in our experiments.

\subsection{Tool Set}
\label{Supply_Tool_use}
For each degradation, we follow previous restoration agents \cite{agenticir, multiAgent} and consider several open-sourced restoration models as the callable tools.
These candidate tools are:
\begin{itemize}[leftmargin=0.5cm]
\item \textit{Brightening}: 
CLAHE \cite{CLAHE},
Gamma correction
($\gamma = 2/3$),
constant shift (adding a constant 40).

\item \textit{Defocus deblurring}: DRBNet \cite{DRBNet},
IFAN \cite{IFAN},
Restormer \cite{Restormer}.

\item \textit{JPEG compression artifact removal}: 
SwinIR \cite{SwinIR} (quality factor 40),
FBCNN \cite{FBCNN} (quality factor 90),
FBCNN \cite{FBCNN} (quality factor 5),
FBCNN \cite{FBCNN} (blind to quality factor).

\item \textit{Denoising}: SwinIR \cite{SwinIR} (noise level 15), SwinIR \cite{SwinIR} (noise level 50), MAXIM \cite{Maxim}, MPRNet \cite{MPRNet}, Restormer \cite{Restormer},
X-Restormer \cite{Xrestormer}.

\item \textit{Deraining}:
MAXIM \cite{Maxim},
MPRNet \cite{MPRNet},
Restormer \cite{Restormer},
X-Restormer \cite{Xrestormer}.

\item \textit{Motion deblurring}: 
MAXIM \cite{Maxim},
MPRNet \cite{MPRNet},
Restormer \cite{Restormer},
X-Restormer \cite{Xrestormer}.

\item \textit{Dehazing}:
MAXIM \cite{Maxim},
X-Restormer \cite{Xrestormer};
RIDCP \cite{RIDCP},
DehazeFormer \cite{DehazeFormer}.
\end{itemize}

\subsection{Evaluation Metrics}
\label{Supply_Metrics}
We assess model performance using three full-reference metrics: PSNR, SSIM \cite{SSIM}, LPIPS \cite{LPIPS}, and four no-reference metrics: MANIQA \cite{MANIQA}, CLIP-IQA \cite{CLIPIQA}, MUSIQ \cite{MUSIQ}, and DeQA-Score \cite{you2025teaching}. 
We briefly introduce these metrics below:
\begin{itemize}[leftmargin=0.5cm] 
    \item \textit{PSNR}: A pixel-level metric that measures the mean squared error between the restored and reference images. Higher PSNR indicates better fidelity.
    \item \textit{SSIM} \cite{SSIM}: Evaluates perceptual similarity by assessing luminance, contrast, and structural components between image pairs. It better aligns with human visual perception than PSNR.
    \item \textit{LPIPS} \cite{LPIPS}: A perceptual metric that uses deep neural network features to compare image similarity, capturing differences that are visually meaningful but not captured by PSNR or SSIM.
    \item \textit{MANIQA} \cite{MANIQA}: A no-reference image quality assessment model that leverages transformer-based architecture and multi-level semantic features to predict perceptual quality without needing a ground-truth image.
    \item \textit{CLIP-IQA} \cite{CLIPIQA}: A no-reference quality assessment model that uses CLIP embeddings to assess image quality based on its alignment with natural image statistics learned from large-scale vision-language pretraining.
    \item \textit{MUSIQ} \cite{MUSIQ}: A transformer-based non-reference image quality assessment model that adapts to various resolutions and content types by processing image patches, offering strong generalization across diverse datasets.

    \item \textit{DeQA-Score} \cite{you2025teaching}:  The model DeQA‑Score computes a continuous image-quality score by first having a MLLM process one or more input images, then outputting a soft-label distribution over discrete quality levels (rather than a simple one-hot label). It treats the human quality ratings as approximately Gaussian and trains the MLLM with a KL-divergence loss to match that soft label. At inference time, the predicted distribution over levels is combined (by a weighted sum over discrete rating tokens) to produce a final quality score that more accurately reflects continuous human judgments.
\end{itemize}

\section{More Results}

\subsection{Qualitative Comparison}
\label{Supply_Quality}
\begin{figure*}
\centering
\includegraphics[scale=0.69,trim=2 0 2 0,clip]{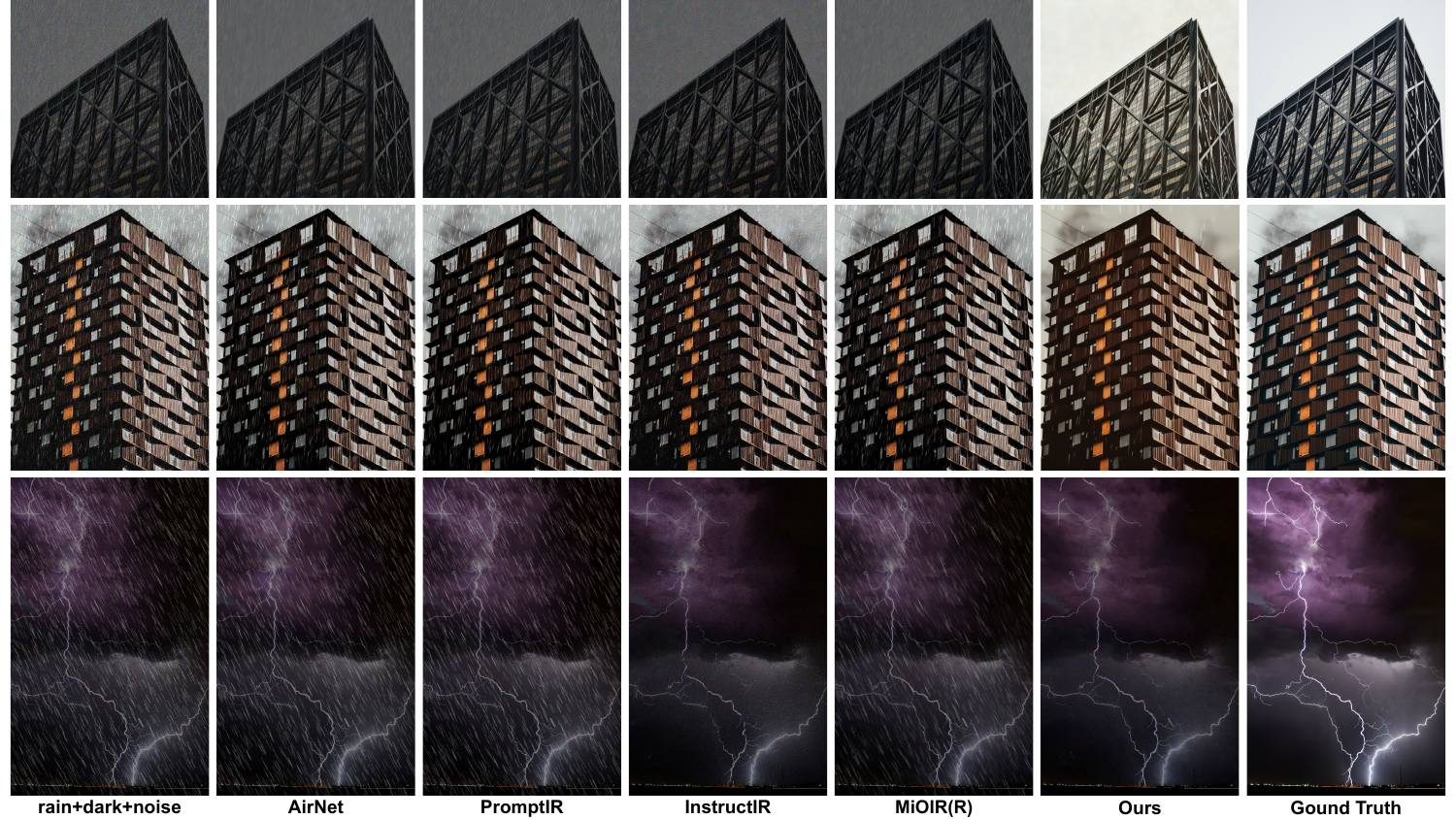}
\includegraphics[scale=0.69,trim=2 0 2 150,clip]{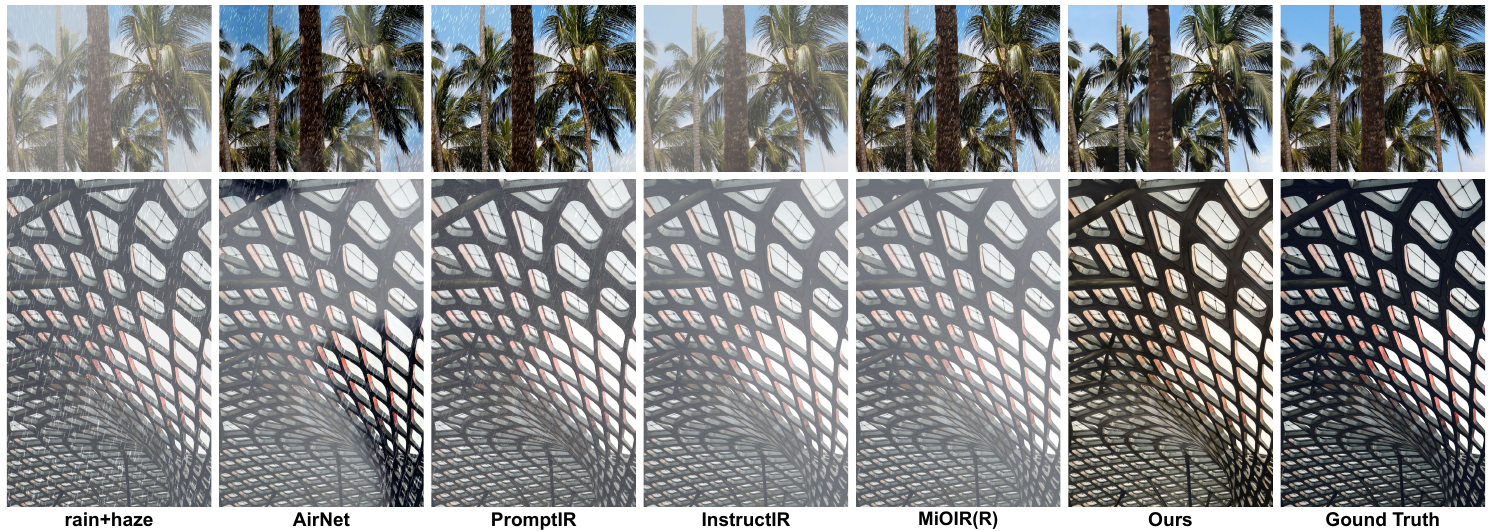}
\caption{Qualitative comparison between our method and SOTA restoration baselines.}
\label{vis_compare_supply}
\end{figure*}

Figure \ref{vis_compare_supply} shows visual comparisons between our method and AirNet \cite{airnet}, PromptIR \cite{Promptir}, InstructIR \cite{Instructir}, and MiOIR(R) \cite{kong2024towards}  under rain+haze and rain+dark+noise degradation cases.
The results further demonstrate that our method effectively removes multiple co-occurring corruptions from degraded images and produces visual quality that is comparable to, or even exceeds, these supervised baselines.

\subsection{Quantitative Comparison}
\label{Supply_Quantitative}
Tabls \ref{supplyd1}, \ref{supplyd2}, \ref{supplyd3} present the performance comparison between our method and the competing baselines across all degradation cases. 
As shown, even without access to ground-truth supervision, our method achieves competitive results on full-reference metrics and consistently outperforms all baselines on no-reference metrics, further demonstrating the effectiveness and robustness of our approach.

\subsection{Perception-Distortion Tradeoff}
Figures \ref{supp_trade_off1} and \ref{supp_trade_off2} illustrate the perception–distortion tradeoff under the defocus blur + JPEG compression and noise + JPEG compression degradation cases. These results lead to the same conclusion as discussed in Section 4.3.

\begin{table*}[!tb]
\centering
\small
\setlength{\tabcolsep}{8.0pt}
\caption{Quantitative comparison across multiple mixed-degradation conditions. Arrows indicate the desired direction of improvement.}
\label{supplyd1}
\begin{tabular}{l|lccccccc}
\toprule
\multirow{2}{*}{} & \multirow{2}{*}{Method} & \multicolumn{3}{>{\columncolor{cyan!10}}c}{Full-Reference}  & \multicolumn{4}{>{\columncolor{green!10}}c}{No-Reference}    \\
&  & \cellcolor{cyan!10}{PSNR}~↑ & \cellcolor{cyan!10}{SSIM}~↑ & \cellcolor{cyan!10}{LPIPS}~↓ & \cellcolor{green!10}{MANIQA}~↑ & \cellcolor{green!10}{CLIP-IQA}~↑ & \cellcolor{green!10}{MUSIQ}~↑  & \cellcolor{green!10}{DeQA-Score)}~↑ \\
\midrule
\multirow{10}{*}{\rotatebox{90}{Case 1}} 
& AirNet \cite{airnet}  &16.272	&0.633	&\cellcolor{cyan!10}0.221	&\cellcolor{green!25}0.423	&0.572	&64.253	&3.539  \\
& PromptIR \cite{Promptir} &16.217	&0.615	&0.242	&\cellcolor{green!10}0.407	&0.540	&\cellcolor{green!10}64.364	&3.479   \\
& InstructIR \cite{Instructir} &15.305	&0.469	&0.496	&0.260	&0.507	&49.016	&3.010   \\
& MiOIR(R) \cite{kong2024towards} &16.563	&0.632	&0.233	&0.312	&\cellcolor{green!10}0.612	&59.093	&3.561  \\
& MiOIR(U) \cite{kong2024towards} &16.696	&\cellcolor{cyan!10}0.675	&\cellcolor{cyan!25}0.210	&0.348	&0.598	&62.262	&\cellcolor{green!10}3.628  \\
& DA-CLIP \cite{daclip} &16.343	&0.552	&0.340	&0.330	&0.494	&57.732	&3.289\\
& AutoDIR \cite{Autodir} &16.771	&0.658	&0.257	&0.362	&0.590	&63.672	&3.625  \\
& AgenticIR \cite{agenticir} &\cellcolor{cyan!25}19.692	&\cellcolor{cyan!25}0.708	&0.338	&0.346	&0.444	&59.594	&3.388   \\
& Ours &\cellcolor{cyan!10}17.995	&0.664	&0.362	&0.399	&\cellcolor{green!25}0.631	&\cellcolor{green!25}68.240	&\cellcolor{green!25}3.871  \\
\midrule
\multirow{10}{*}{\rotatebox{90}{Case 2}} 
& AirNet \cite{airnet} &22.282	&0.629	&0.501	&0.194	&0.232	&28.489	&2.282 \\
& PromptIR \cite{Promptir} &23.110	&0.632	&0.486	&0.203	&0.247	&28.715	&2.298  \\
& InstructIR \cite{Instructir} &23.727	&0.648	&\cellcolor{cyan!25}0.519	&0.211	&0.216	&29.441	&2.324  \\
& MiOIR(R) \cite{kong2024towards} &\cellcolor{cyan!10}23.960	&0.653	&0.503	&0.209	&0.256	&30.628	&2.353  \\
& MiOIR(U) \cite{kong2024towards} &\cellcolor{cyan!25}23.965	&0.653	&\cellcolor{cyan!10}0.511	&0.203	&0.278	&30.848	&2.338   \\
& DA-CLIP \cite{daclip} &23.580	&0.647	&0.490	&0.207	&0.272	&30.073	&2.354 \\
& AutoDIR \cite{Autodir} &23.915	&\cellcolor{cyan!10}0.654	&0.439	&0.215	&0.269	&36.215	&2.571  \\
& AgenticIR \cite{agenticir} &22.755	&\cellcolor{cyan!25}0.658	&0.435	&\cellcolor{green!10}0.216	&\cellcolor{green!25}0.303	&\cellcolor{green!10}43.866	&\cellcolor{green!10}2.688  \\
& Ours &22.338	&0.634	&0.456	&\cellcolor{green!25}0.225	&\cellcolor{green!10}0.302	&\cellcolor{green!25}44.950	&\cellcolor{green!25}2.832   \\
\midrule
\multirow{10}{*}{\rotatebox{90}{Case 3}}
& AirNet \cite{airnet} &14.083	&0.450	&0.411	&0.175	&0.281	&42.572	&2.429 \\
& PromptIR \cite{Promptir} &14.061	&0.433	&0.406	&0.183	&0.290	&42.710	&2.434   \\
& InstructIR \cite{Instructir} &14.489	&0.461	&0.413	&0.182	&0.278	&42.521	&2.481 \\
& MiOIR(R) \cite{kong2024towards} &\cellcolor{cyan!10}17.294	&0.561	&0.371	&0.176	&0.281	&43.473	&2.497 \\
& MiOIR(U) \cite{kong2024towards} &16.989	&\cellcolor{cyan!25}0.574	&0.388	&0.170	&0.276	&43.659	&2.471  \\
& DA-CLIP \cite{daclip} &15.520	&0.517	&0.356	&0.201	&0.299	&48.516	&2.728  \\
& AutoDIR \cite{Autodir} &15.984	&\cellcolor{cyan!10}0.567	&\cellcolor{cyan!10}0.319	&0.254	&0.364	&56.248	&\cellcolor{green!10}3.340  \\
& AgenticIR \cite{agenticir} &\cellcolor{cyan!25}17.306	&0.519	&\cellcolor{cyan!25}0.314	&\cellcolor{green!10}0.269	&\cellcolor{green!10}0.406	&\cellcolor{green!10}59.304	&3.325  \\
& Ours &15.691	&0.482	&0.393	&\cellcolor{green!25}0.345	&\cellcolor{green!25}0.582	&\cellcolor{green!25}66.178	&\cellcolor{green!25}3.797  \\
\midrule
\multirow{10}{*}{\rotatebox{90}{Case 4}}
& AirNet \cite{airnet} &23.522	&0.613	&0.458	&0.287	&0.379	&44.700	&2.958  \\
& PromptIR \cite{Promptir} &24.249	&0.620	&0.448	&0.293	&0.408	&45.182	&2.982 \\
& InstructIR \cite{Instructir} &24.238	&0.626	&0.439	&0.294	&0.413	&45.629	&3.028  \\
& MiOIR(R) \cite{kong2024towards} &25.453	&0.671	&0.381	&0.320	&0.464	&52.704	&3.137   \\
& MiOIR(U) \cite{kong2024towards} &\cellcolor{cyan!25}25.628	&0.673	&0.373	&0.324	&0.486	&53.445	&3.086  \\
& DA-CLIP \cite{daclip} &24.592	&0.632	&0.403	&0.316	&\cellcolor{green!25}0.506	&49.354	&3.123 \\
& AutoDIR \cite{Autodir} &24.221	&0.622	&0.436	&0.295	&0.437	&45.982	&3.030 \\
& AgenticIR \cite{agenticir} &\cellcolor{cyan!10}25.489	&\cellcolor{cyan!25}0.800	&\cellcolor{cyan!25}0.258	&\cellcolor{green!10}0.344	&0.462	&\cellcolor{green!10}61.722	&\cellcolor{green!10}3.720 \\
& Ours &24.834	&\cellcolor{cyan!10}0.770	&\cellcolor{cyan!10}0.281	&\cellcolor{green!25}0.374	&\cellcolor{green!10}0.487	&\cellcolor{green!25}63.852	&\cellcolor{green!25}3.774  \\
\midrule
\multirow{10}{*}{\rotatebox{90}{Case 5}}
& AirNet \cite{airnet} &16.288	&0.716	&0.279	&0.396	&0.531	&64.235	&3.349  \\
& PromptIR \cite{Promptir} &\cellcolor{cyan!10}18.351	&0.745	&0.275	&\cellcolor{green!25}0.414	&0.515	&64.975	&3.341 \\
& InstructIR \cite{Instructir} &14.213	&0.727	&0.215	&0.392	&0.538	&63.626	&3.377  \\
& MiOIR(R) \cite{kong2024towards} &16.578	&0.776	&0.197	&0.395	&0.539	&65.445	&3.444  \\
& MiOIR(U) \cite{kong2024towards} &16.322	&\cellcolor{cyan!10}0.784	&\cellcolor{cyan!25}0.177	&0.401	&0.541	&65.869	&3.533  \\
& DA-CLIP \cite{daclip} &16.225	&0.722	&0.262	&\cellcolor{green!10}0.408	&0.515	&65.923	&3.374 \\
& AutoDIR \cite{Autodir} &16.136	&0.721	&0.280	&0.402	&0.497	&65.782	&3.365 \\
& AgenticIR \cite{agenticir} &\cellcolor{cyan!25}19.373	&\cellcolor{cyan!25}0.805	&\cellcolor{cyan!10}0.184	&0.400	&\cellcolor{green!10}0.592	&\cellcolor{green!10}68.463	&\cellcolor{green!10}3.994 \\
& Ours &16.705	&0.687	&0.333	&0.403	&\cellcolor{green!25}0.624	&\cellcolor{green!25}71.796	&\cellcolor{green!25}4.010 \\
\bottomrule
\end{tabular} 
\end{table*}

\begin{table*}[!tb]
\centering
\small
\setlength{\tabcolsep}{8.0pt}
\caption{Quantitative comparison across multiple mixed-degradation conditions. Arrows indicate the desired direction of improvement.}
\label{supplyd2}
\begin{tabular}{l|lccccccc}
\toprule
\multirow{2}{*}{} & \multirow{2}{*}{Method} & \multicolumn{3}{>{\columncolor{cyan!10}}c}{Full-Reference}  & \multicolumn{4}{>{\columncolor{green!10}}c}{No-Reference}    \\
&  & \cellcolor{cyan!10}{PSNR}~↑ & \cellcolor{cyan!10}{SSIM}~↑ & \cellcolor{cyan!10}{LPIPS}~↓ & \cellcolor{green!10}{MANIQA}~↑ & \cellcolor{green!10}{CLIP-IQA}~↑ & \cellcolor{green!10}{MUSIQ}~↑  & \cellcolor{green!10}{DeQA-Score)}~↑ \\
\midrule
\multirow{10}{*}{\rotatebox{90}{Case 6}} 
& AirNet \cite{airnet}  &14.781	&0.657	&0.313	&\cellcolor{green!10}0.373	&0.492	&\cellcolor{green!10}63.081	&3.293 \\
& PromptIR \cite{Promptir} &14.783	&0.661	&0.310	&0.355	&0.467	&62.443	&3.320  \\
& InstructIR \cite{Instructir} &14.456	&0.446	&0.646	&0.248	&0.469	&43.749	&2.770   \\
& MiOIR(R) \cite{kong2024towards} &14.672	&0.632	&0.320	&0.267	&\cellcolor{green!10}0.562	&55.223	&3.390  \\
& MiOIR(U) \cite{kong2024towards} &14.830	&0.665	&\cellcolor{cyan!10}0.285	&0.302	&0.487	&60.079	&3.361   \\
& DA-CLIP \cite{daclip} &14.503	&0.602	&0.397	&0.313	&0.399	&56.516	&3.134  \\
& AutoDIR \cite{Autodir} &\cellcolor{cyan!10}17.249	&\cellcolor{cyan!10}0.699	&\cellcolor{cyan!25}0.268	&0.322	&0.518	&62.531	&3.567   \\
& AgenticIR \cite{agenticir} &\cellcolor{cyan!25}18.725	&\cellcolor{cyan!25}0.699	&0.349	&0.315	&0.433	&59.596	&\cellcolor{green!10}3.580   \\
& Ours &15.999	&0.623	&0.389	&\cellcolor{green!25}0.393	&\cellcolor{green!25}0.623	&\cellcolor{green!25}68.938	&\cellcolor{green!25}3.869   \\
\midrule
\multirow{10}{*}{\rotatebox{90}{Case 7}} 
& AirNet \cite{airnet} &20.800	&0.623	&0.401	&0.198	&0.249	&34.574	&2.591 \\
& PromptIR \cite{Promptir} &21.494	&0.632	&0.385	&0.207	&0.271	&34.707	&2.630  \\
& InstructIR \cite{Instructir} &21.738	&0.640	&0.400	&0.210	&0.255	&36.812	&2.665  \\
& MiOIR(R) \cite{kong2024towards} &\cellcolor{cyan!10}21.941	&\cellcolor{cyan!10}0.648	&0.389	&0.193	&0.302	&39.335	&2.640  \\
& MiOIR(U) \cite{kong2024towards} &\cellcolor{cyan!25}21.949	&\cellcolor{cyan!25}0.648	&0.393	&0.186	&0.326	&39.704	&2.619   \\
& DA-CLIP \cite{daclip} &21.664	&0.640	&0.384	&0.193	&0.295	&38.169	&2.641  \\
& AutoDIR \cite{Autodir} &21.679	&0.634	&\cellcolor{cyan!25}0.374	&\cellcolor{green!10}0.222	&0.296	&38.146	&2.822  \\
& AgenticIR \cite{agenticir} &20.397	&0.625	&\cellcolor{cyan!10}0.383	&0.212	&\cellcolor{green!10}0.349	&\cellcolor{green!10}48.192	&\cellcolor{green!10}2.922  \\
& Ours &20.370	&0.603	&0.411	&\cellcolor{green!25}0.245	&\cellcolor{green!25}0.350	&\cellcolor{green!25}50.361	&\cellcolor{green!25}3.061  \\
\midrule
\multirow{10}{*}{\rotatebox{90}{Case 8}}
& AirNet \cite{airnet} &16.006	&0.647	&0.263	&0.399	&0.576	&63.726	&3.409   \\
& PromptIR \cite{Promptir} &16.178	&0.687	&0.207	&\cellcolor{green!10}0.411	&0.585	&64.320	&3.512  \\
& InstructIR \cite{Instructir} &16.329	&0.692	&0.190	&0.400	&0.574	&63.881	&3.548  \\
& MiOIR(R) \cite{kong2024towards} &\cellcolor{cyan!10}17.522	&\cellcolor{cyan!10}0.732	&0.202	&0.404	&0.573	&65.350	&3.489  \\
& MiOIR(U) \cite{kong2024towards} &16.580	&0.715	&\cellcolor{cyan!25}0.156	&0.409	&\cellcolor{green!10}0.585	&65.647	&3.592  \\
& DA-CLIP \cite{daclip} &16.917	&0.669	&0.266	&\cellcolor{green!25}0.419	&0.548	&66.502	&3.438 \\
& AutoDIR \cite{Autodir} &16.691	&0.660	&0.314	&0.410	&0.530	&66.372	&3.382 \\
& AgenticIR \cite{agenticir} &\cellcolor{cyan!25}22.132	&\cellcolor{cyan!25}0.832	&\cellcolor{cyan!10}0.169	&0.402	&0.561	&\cellcolor{green!10}68.141	&\cellcolor{green!10}3.800  \\
& Ours &17.506	&0.682	&0.356	&0.390	&\cellcolor{green!25}0.618	&\cellcolor{green!25}70.732	&\cellcolor{green!25}3.866   \\
\midrule
\multirow{10}{*}{\rotatebox{90}{Case 9}}
& AirNet \cite{airnet} &14.814	&0.458	&0.544	&0.189	&0.192	&28.382	&2.131  \\
& PromptIR \cite{Promptir} &15.014	&0.446	&0.534	&0.196	&0.202	&28.532	&2.128 \\
& InstructIR \cite{Instructir} &15.434	&0.483	&0.550	&0.203	&0.182	&29.241	&2.165  \\
& MiOIR(R) \cite{kong2024towards} &15.491	&0.490	&0.550	&0.206	&0.233	&30.287	&2.197  \\
& MiOIR(U) \cite{kong2024towards} &15.483	&0.489	&0.555	&0.201	&0.260	&30.538	&2.184   \\
& DA-CLIP \cite{daclip} &15.778	&0.496	&0.543	&\cellcolor{green!10}0.206	&0.244	&30.166	&2.224 \\
& AutoDIR \cite{Autodir} &15.973	&0.514	&\cellcolor{cyan!10}0.488	&0.196	&0.219	&34.427	&2.338  \\
& AgenticIR \cite{agenticir} &\cellcolor{cyan!25}18.798	&\cellcolor{cyan!25}0.584	&0.495	&0.196	&\cellcolor{green!10}0.262	&\cellcolor{green!10}41.299	&\cellcolor{green!10}2.504   \\
& Ours &\cellcolor{cyan!10}16.287	&\cellcolor{cyan!10}0.535	&\cellcolor{cyan!25}0.482	&\cellcolor{green!25}0.304	&\cellcolor{green!25}0.491	&\cellcolor{green!25}60.227	&\cellcolor{green!25}3.438   \\
\midrule
\multirow{10}{*}{\rotatebox{90}{Case 10}}
& AirNet \cite{airnet} &\cellcolor{cyan!10}21.250	&0.559	&0.536	&0.196	&0.277	&29.854	&2.096 \\
& PromptIR \cite{Promptir} &\cellcolor{cyan!25}21.269	&\cellcolor{cyan!10}0.562	&0.562	&\cellcolor{green!10}0.196	&0.254	&28.904	&2.086   \\
& InstructIR \cite{Instructir} &19.104	&0.347	&0.780	&0.103	&0.305	&22.685	&1.882 \\
& MiOIR(R) \cite{kong2024towards} &20.923	&0.530	&0.569	&0.116	&\cellcolor{green!25}0.345	&25.380	&2.034  \\
& MiOIR(U) \cite{kong2024towards} &21.225	&\cellcolor{cyan!25}0.566	&0.526	&0.153	&0.278	&26.593	&2.090  \\
& DA-CLIP \cite{daclip} &20.474	&0.486	&0.624	&0.163	&0.295	&25.290	&2.019  \\
& AutoDIR \cite{Autodir} &18.855	&0.479	&\cellcolor{cyan!25}0.520	&\cellcolor{green!25}0.241	&\cellcolor{green!10}0.335	&\cellcolor{green!25}41.623	&\cellcolor{green!25}2.528 \\
& AgenticIR \cite{agenticir} &20.034	&0.546	&\cellcolor{cyan!10}0.525	&0.195	&0.292	&\cellcolor{green!10}38.196	&\cellcolor{green!10}2.401 \\
& Ours &20.202	&0.530	&0.531	&0.186	&0.288	&37.379	&2.368   \\
\bottomrule
\end{tabular} 
\end{table*}

\begin{table*}[!tb]
\centering
\small
\setlength{\tabcolsep}{8.0pt}
\caption{Quantitative comparison across multiple mixed-degradation conditions. Arrows indicate the desired direction of improvement.}
\label{supplyd3}
\begin{tabular}{l|lccccccc}
\toprule
\multirow{2}{*}{} & \multirow{2}{*}{Method} & \multicolumn{3}{>{\columncolor{cyan!10}}c}{Full-Reference}  & \multicolumn{4}{>{\columncolor{green!10}}c}{No-Reference}    \\
&  & \cellcolor{cyan!10}{PSNR}~↑ & \cellcolor{cyan!10}{SSIM}~↑ & \cellcolor{cyan!10}{LPIPS}~↓ & \cellcolor{green!10}{MANIQA}~↑ & \cellcolor{green!10}{CLIP-IQA}~↑ & \cellcolor{green!10}{MUSIQ}~↑  & \cellcolor{green!10}{DeQA-Score)}~↑ \\
\midrule
\multirow{10}{*}{\rotatebox{90}{Case 11}} 
& AirNet \cite{airnet}  &16.650	&0.521	&0.452	&\cellcolor{green!25}0.371	&0.499	&\cellcolor{green!10}62.832	&3.060 \\
& PromptIR \cite{Promptir} &16.601	&0.511	&0.460	&\cellcolor{green!10}0.354	&0.493	&62.143	&3.065  \\
& InstructIR \cite{Instructir} &15.790	&0.455	&0.560	&0.236	&0.455	&47.557	&2.826   \\
& MiOIR(R) \cite{kong2024towards} &16.734	&0.529	&0.438	&0.285	&0.525	&56.743	&3.200   \\
& MiOIR(U) \cite{kong2024towards} &16.487	&0.532	&0.451	&0.319	&0.537	&60.476	&3.142   \\
& DA-CLIP \cite{daclip} &16.241	&0.443	&0.514	&0.296	&0.443	&55.044	&2.971 \\
& AutoDIR \cite{Autodir} &16.776	&0.575	&\cellcolor{cyan!25}0.387	&0.324	&\cellcolor{green!10}0.554	&62.473	&\cellcolor{green!10}3.373 \\
& AgenticIR \cite{agenticir} &\cellcolor{cyan!25}18.674	&\cellcolor{cyan!25}0.661	&\cellcolor{cyan!10}0.404	&0.287	&0.394	&57.342	&3.254  \\
& Ours &\cellcolor{cyan!10}17.464	&\cellcolor{cyan!10}0.596	&0.440	&0.343	&\cellcolor{green!25}0.569	&\cellcolor{green!25}66.904	&\cellcolor{green!25}3.582  \\
\midrule
\multirow{10}{*}{\rotatebox{90}{Case 12}} 
& AirNet \cite{airnet} &13.162	&0.528	&0.483	&\cellcolor{green!10}0.327	&0.429	&\cellcolor{green!10}61.533	&2.871   \\
& PromptIR \cite{Promptir} &13.167	&0.523	&0.502	&0.300	&0.422	&59.385	&2.866  \\
& InstructIR \cite{Instructir} &13.320	&0.370	&0.767	&0.227	&0.428	&39.365	&2.451  \\
& MiOIR(R) \cite{kong2024towards} &13.151	&0.508	&0.490	&0.252	&0.474	&52.839	&3.000   \\
& MiOIR(U) \cite{kong2024towards} &13.202	&0.527	&0.489	&0.273	&0.430	&57.886	&2.927  \\
& DA-CLIP \cite{daclip} &13.058	&0.453	&0.567	&0.265	&0.386	&53.851	&2.739\\
& AutoDIR \cite{Autodir} &15.875	&\cellcolor{cyan!10}0.592	&\cellcolor{cyan!25}0.410	&0.292	&\cellcolor{green!10}0.474	&60.303	&\cellcolor{green!10}3.228 \\
& AgenticIR \cite{agenticir} &\cellcolor{cyan!25}16.895	&\cellcolor{cyan!25}0.612	&\cellcolor{cyan!10}0.435	&0.263	&0.399	&54.407	&3.184  \\
& Ours &\cellcolor{cyan!10}16.648	&0.591	&0.447	&\cellcolor{green!25}0.345	&\cellcolor{green!25}0.542	&\cellcolor{green!25}66.515	&\cellcolor{green!25}3.600  \\
\midrule
\multirow{10}{*}{\rotatebox{90}{Case 13}}
& AirNet \cite{airnet} &14.032	&0.467	&0.542	&0.174	&0.183	&30.106	&2.185  \\
& PromptIR \cite{Promptir} &\cellcolor{cyan!10}15.082	&0.506	&0.521	&0.191	&0.194	&30.544	&2.237  \\
& InstructIR \cite{Instructir} &14.738	&0.508	&0.548	&0.199	&0.177	&32.111	&2.264   \\
& MiOIR(R) \cite{kong2024towards} &14.787	&0.513	&0.544	&0.187	&0.245	&34.559	&2.285 \\
& MiOIR(U) \cite{kong2024towards} &14.769	&\cellcolor{cyan!10}0.513	&0.551	&0.178	&0.272	&35.271	&2.263  \\
& DA-CLIP \cite{daclip} &14.590	&0.505	&0.539	&0.192	&0.231	&33.721	&2.322\\
& AutoDIR \cite{Autodir} &14.821	&0.509	&0.525	&\cellcolor{green!10}0.207	&0.216	&33.228	&2.372 \\
& AgenticIR \cite{agenticir} &\cellcolor{cyan!25}15.514	&\cellcolor{cyan!25}0.520	&\cellcolor{cyan!25}0.513	&0.181	&\cellcolor{green!10}0.277	&\cellcolor{green!10}41.847	&\cellcolor{green!10}2.405  \\
& Ours &14.671	&0.476	&\cellcolor{cyan!10}0.517	&\cellcolor{green!25}0.276	&\cellcolor{green!25}0.448	&\cellcolor{green!25}56.298	&\cellcolor{green!25}3.168   \\
\midrule
\multirow{10}{*}{\rotatebox{90}{Case 14}}
& AirNet \cite{airnet} &\cellcolor{cyan!10}15.499	&0.465	&0.635	&0.152	&0.195	&23.437	&1.842  \\
& PromptIR \cite{Promptir} &15.397	&\cellcolor{cyan!10}0.475	&0.634	&0.171	&0.207	&23.194	&1.924  \\
& InstructIR \cite{Instructir} &12.947	&0.452	&0.682	&0.186	&0.195	&23.107	&1.937   \\
& MiOIR(R) \cite{kong2024towards} &12.791	&0.449	&0.693	&0.188	&0.213	&23.952	&1.967  \\
& MiOIR(U) \cite{kong2024towards} &12.775	&0.450	&0.704	&0.185	&\cellcolor{green!10}0.241	&24.009	&1.967  \\
& DA-CLIP \cite{daclip} &12.772	&0.442	&0.689	&0.194	&0.239	&24.610	&1.967  \\
& AutoDIR \cite{Autodir} &13.324	&0.428	&0.645	&\cellcolor{green!10}0.205	&0.229	&30.584	&2.069  \\
& AgenticIR \cite{agenticir} &\cellcolor{cyan!25}16.214	&\cellcolor{cyan!25}0.491	&\cellcolor{cyan!10}0.602	&0.159	&0.250	&\cellcolor{green!10}35.394	&\cellcolor{green!10}2.175  \\
& Ours &15.120	&0.447	&\cellcolor{cyan!25}0.570	&\cellcolor{green!25}0.266	&\cellcolor{green!25}0.472	&\cellcolor{green!25}57.365	&\cellcolor{green!25}3.217  \\
\midrule
\multirow{10}{*}{\rotatebox{90}{Case 15}}
& AirNet \cite{airnet} &18.854	&0.484	&0.640	&0.138	&0.213	&20.739	&1.797  \\
& PromptIR \cite{Promptir} &\cellcolor{cyan!25}19.598	&0.495	&0.627	&0.141	&0.234	&20.752	&1.800  \\
& InstructIR \cite{Instructir} &\cellcolor{cyan!10}19.420	&\cellcolor{cyan!25}0.504	&0.667	&0.150	&0.204	&20.900	&1.823  \\
& MiOIR(R) \cite{kong2024towards} &19.245	&0.500	&0.662	&0.159	&0.218	&21.385	&1.838   \\
& MiOIR(U) \cite{kong2024towards} &19.244	&\cellcolor{cyan!10}0.502	&0.674	&\cellcolor{green!10}0.159	&0.248	&21.425	&1.835  \\
& DA-CLIP \cite{daclip} &19.110	&0.498	&0.649	&0.158	&0.237	&22.256	&1.854  \\
& AutoDIR \cite{Autodir} &18.808	&0.456	&\cellcolor{cyan!25}0.576	&\cellcolor{green!25}0.196	&0.248	&32.703	&\cellcolor{green!10}2.155  \\
& AgenticIR \cite{agenticir} &18.910	&0.484	&\cellcolor{cyan!10}0.604	&0.156	&\cellcolor{green!25}0.250	&\cellcolor{green!10}34.060	&2.042  \\
& Ours &18.750	&0.497	&0.605	&0.156	&\cellcolor{green!10}0.248	&\cellcolor{green!25}35.768	&\cellcolor{green!25}2.207   \\
\bottomrule
\end{tabular} 
\end{table*}